\RequirePackage{fix-cm}
\documentclass[smallcondensed]{svjour3}     
\smartqed  
\usepackage{amsmath}
\usepackage[utf8]{inputenc}
\usepackage{graphicx}
\usepackage{amsfonts,amssymb}
\usepackage{stackrel}
\usepackage{natbib}
\usepackage{xcolor}
\usepackage{color}
\usepackage[ruled,vlined]{algorithm2e}

\usepackage{mathptmx}      
\usepackage{latexsym}
\journalname{Statistics and Computing}
\begin{document}

\title{Accelerated Parallel Non-conjugate Sampling for Bayesian Non-parametric Models
\thanks{The contribution of Michael Zhang and Sinead Williamson was funded by NSF grant IIS 1447721. The contribution of Michael Zhang was also funded by the University of Hong Kong's Seed Fund for Basic Research for New Staff.}
}


\author{Michael Minyi Zhang         \and
        Sinead A. Williamson \and
         Fernando P\'{e}rez-Cruz 
}


\institute{M. M. Zhang \at
            Department of Statistics and Actuarial Science\\ 
            University of Hong Kong\\
            \email{mzhang18@hku.hk}           
           \and
           S. A. Williamson \at
             Department of Statistics and Data Science\\ 
             The University of Texas at Austin\\
             \email{sinead.williamson@mccombs.utexas.edu}
            \and
            F. P\'{e}rez-Cruz \at
            Swiss Data Science Center\\
            ETH Z\"{u}rich \\
            \email{fernando.perezcruz@sdsc.ethz.ch}
}

\date{Received: \today / Accepted: date}

\maketitle

\begin{abstract}
	Inference of latent feature models in the Bayesian nonparametric setting is generally difficult, especially in high dimensional settings, because it usually requires proposing features from some prior distribution. In special cases, where the integration is tractable, we can sample new feature assignments according to a predictive likelihood. 
We present a novel method to accelerate the mixing of latent variable model inference by proposing feature locations based on the data, as opposed to the prior. First, we introduce an accelerated feature proposal mechanism that we show is a valid MCMC algorithm for posterior inference. Next, we propose an approximate inference strategy to perform accelerated inference in parallel. A two-stage algorithm that combines the two approaches provides a computationally attractive method that can quickly reach local convergence to the posterior distribution of our model, while allowing us to exploit parallelization.


\keywords{Machine learning \and Bayesian Non-parametrics \and Scalable inference \and Parallel computing}

\end{abstract}

\section{Introduction}\label{sec:intro}
Bayesian non-parametric (BNP) models appear to be perfectly suited for the era of big data \citep{Jordan:2011}, in which ever-expanding databases of high-dimensional data cannot be dealt with simplistically. Discrete non-parametric priors like the Dirichlet process \citep{Ferguson:1973} or the beta-Bernoulli process \citep{Griffiths:Ghahramani:2011} allow for modeling latent variables like clusters or otherwise unobservable features in our data and adapting the complexity of the model in accordance to the complexity of the data. Even if we had some understanding of the latent structure in the data, we would not necessarily know their exact forms and implications in the model \textit{a priori}. The BNP solution, which divides the data into discrete features or clusters, fosters interpretable models that would naturally lead to new hypotheses about the information in such databases \citep{Been:2015}. For example, in a general medical records data set containing billions of observations, a cluster (or feature) composed of 0.001\% of the population still includes tens of thousands of people. But when looking at a small fraction of the data, these clusters may be seen as outliers at best and it would be difficult to characterize their meaning. Exploring large, complicated databases with the intention of finding interpretable new features by adapting the model complexity to the complexity of the data is BNP's most prominent and promising feature. Unfortunately, BNP fails to deliver on this promise due to at least three limitations:

\begin{enumerate}
	\item \textbf{Base measure definition}: The prior distribution over the parameters associated with a latent feature is governed by a (typically diffuse) base measure, which divides the probability mass over the entire parameter space associated with our predefined likelihood model. In a non-parametric setting, where we typically assume an unbounded number of distinct features, it is generally hard to specify informative priors, meaning the base measure behaves more as a regularizer than a source of prior knowledge about the problem at hand \citep{Gelman:Shalizi:2013}. In a high-dimensional parameter space, this means that, in the absence of highly informative priors, the prior probability mass around the true parameter will be vanishingly small.  
	\item \textbf{Computational complexity}: Inference in discrete BNP models where we are learning feature allocations is a daunting task because it involves exploring a countable, infinite-dimensional state space. This has two consequences for designing scalable inference. First, alignment issues make parallelization difficult. In a parametric setting, we know we have at most $K$ distinct features. Therefore, if all processors exhibit $K$ features, there must be a one-to-one mapping of these features across the processors. However, in a BNP setting, we must align an unbounded number of features across all processors, with no guarantees on the number of features shared between processors.  Features on one processor may not have direct counterparts on another processor---meaning we must decide both \textit{whether} a feature has a counterpart on another processor, and if so \textit{which} feature it corresponds to. To combat this alignment issues, many distributed BNP algorithms rely on approximate matching techniques such as the Hungarian algorithm to align features from different processors \citep{Newman:Asuncion:Smyth:Welling:2009}.

	

Second, exploring the infinite-dimensional space associated with BNP models requires proposing previously unseen features. As we discussed above, in high dimensional spaces the prior probability mass near ``good'' features can be vanishingly small, meaning such exploration can take an unfeasibly long time which is unacceptable under finite computing time constraints. 	
	\item \textbf{Shallow models}: A Bayesian model is defined by a distribution over parameter space (the prior, or later the posterior), and a mapping from parameter space to observation space (the likelihood model). BNP models provide very rich priors. However, in most applications they are combined with a fairly simple likelihood, such as a mixture of Gaussians. These simple likelihoods limit the type of information that can be represented, and cannot capture high-level, complex features in image or audio. By contrast, deep learning-based methods such as convolutional neural networks, recurrent neural networks, and generative adversarial networks are capable of learning complex representations of image and audio data. However, such approaches lack interpretability. Further, performing full posterior inference in such deep models is generally ineffective \citep[see][]{Papamarkou:Hinkle:Young:2019}. We note that the use of  simple likelihoods is common across both non-parametric and parametric Bayesian models, although the added computational complexity of BNP posterior inference makes them perhaps more prevalent in this context. Whether working in a parametric or a non-parametric setting, we need more sophisticated likelihood models to be able to capture the the inherently complex nature of image, audio and video data.
\end{enumerate}


These two limitations are in a sense intrinsic to the BNP setting: while we focus on the Dirichlet process, they are inherent in models that involve a countably infinite parameter space, such as the beta-Bernoulli process. Conversely, they are less problematic in finite-dimensional models. The third limitation, conversely, is a more general limitation of the Bayesian modeling paradigm. There are many lines of research aiming to address the problem of shallow likelihood models, including variational autoencoders \citep{kingma:2014}, adversarial variational Bayes \citep{Mescheder:2017} or deep hierarchical implicit models \citep{Tran:2017}. While the problem of shallow likelihood models is certainly relevant in the BNP setting, incorporating these approaches is not straightforward in our current inference strategy.

In this paper, propose a simple yet powerful method to address the first two limitations, with the goal of improving inference for BNP models in terms of speed and mixing quality, while still maintaining a correct algorithm. We apply our inference algorithm to arguably the most popular BNP-based model---the Dirichlet process mixture model. 

The central idea of our paper is a data-driven adaptation of a popular MCMC algorithm for the Dirichlet process mixture model \citep{Neal:1998}. To improve convergence, instead of sampling from the base measure, we propose a simple mechanism that samples from the data points that are not well characterized by their current clusters. This is augmented with a parallelizable procedure in which, at regular intervals, the different computation nodes share summary statistics and newly introduced features.  With this method, we are able to find new clusters efficiently without needing to sample from the base measure.

We begin this paper with a brief introduction to the Dirichlet processes and inference for Dirichlet process mixture models in Section~\ref{sec:background}. Next in Section~\ref{sec:method}, we present our novel accelerated inference algorithm, first as an exact MCMC method and then as an approximate parallelizable algorithm. We then illustrate the performance of our algorithm in Section~\ref{sec:experiments} with high dimensional, high signal to noise ratio data in which standard MCMC inference procedures for BNP models will fail to find any meaningful information and our algorithm is able to find relevant clusters. And lastly, we conclude our paper in Section~\ref{sec:discussion} with a discussion of our work and present future work for this idea.

\section{Background}\label{sec:background}
Most Bayesian non-parametric models involve placing an infinite dimensional prior on a latent variable space. The inference challenge then becomes inferring the posterior distribution over this latent space, which can be seen as a latent representation of our data. In this paper, we focus on mixture models, where each observation is associated with a single latent variable. The most common non-parametric prior in this context is the Dirichlet process mixture model \citep[DPMM,][]{Antoniak:1974}, and we choose to make this the focus of our investigation. However, our proposed method could be adapted to a wider variety of non-parametric mixture models.

\subsection{The Dirichlet process}
A Dirichlet process (DP) is a distribution over probability measures $G\sim \mbox{DP}(\alpha, H)$, defined on some parameter space, $\Theta$. Here, $H$ is a probability measure on $\Theta$ and $\alpha>0$ is known as the concentration parameter. The Dirichlet process is uniquely defined by its finite-dimensional marginals: if $G\sim \mbox{DP}(\alpha, H)$, then the masses $G(A_1), \dots, G(A_k)$ assigned to any finite partition $A_1,\dots, A_k$ of $\Theta$ are $\mbox{Dirichlet}\left(\alpha H(A_1), \dots, \alpha H(A_k)\right)$-distributed \citep{Ferguson:1973}. The resulting measure $G$ is discrete almost surely.

We can use the Dirichlet process to construct a hierarchical distribution known as a Dirichlet process mixture model, where  
\begin{equation}
\begin{split}
G| \alpha , H &\sim  \mbox{DP}(\alpha,H)\\
\phi_{i} | G &\sim G\\
x_i | \phi_{i} &\sim f(x_i|\phi_{i}).
\end{split}\label{eqn:dpmm_1}
\end{equation}
Here, $\mathcal{F} = \{f(\cdot|\theta), \theta\in \Theta)\}$ is a family of parametric likelihood functions. We will refer to each mixture component $f(\cdot|\phi_i)$ as a \textit{feature}, with feature parameter $\phi_i\in \Theta$. The discrete nature of $G$ means that each feature may be selected multiple times, leading to clustering behavior. 

To make this clustering more explicit, we can separate the random measure $G$ into an infinite sequence of weights $\pi = (\pi_1,\pi_2,\dots)$, and a corresponding infinite sequence of feature parameters $\theta = (\theta_1,\theta_2,\dots)$, where $\theta_k\stackrel{\small{iid}}{\sim}H$. The distribution over weights $\pi$, under a size-biased random re-ordering, is described by the following ``stick breaking'' construction, $\pi \sim \text{\sc{Stick}}(\alpha)$ \citep{Sethuraman:1994}:
\begin{equation}
\begin{split}
\beta_k &\sim \mbox{Beta}(1, \alpha)\\
\pi_k &= \beta_k \prod_{k^{\prime}=1}^{k-1}(1-\beta_{k^{\prime}})
\end{split}
\end{equation}

Decomposing $G$ in this manner allows us to introduce feature indicator variables $z_i$: 
\begin{equation}
    \begin{split}
        \Pr(z_i = k) =& \pi_k\\
        x_i|z_i \sim& f(x_i|\theta_{z_i}).
    \end{split}
\end{equation}

Rather than instantiate the vector of feature probabilities $\pi$, we can integrate out this infinite object, yielding an exchangeable distribution over cluster indicator variables \citep{Aldous:1983}, where
$$Pr(z_i=k|z_{<i}) \propto \begin{cases}\sum_{j<i} \mathbb{I}(z_j=z_i) & k\in K_i\\ \alpha & k\not \in K_i\end{cases}$$
where $\mathbb{I}(\cdot)$ is an indicator function, and $K_i$ is the set of indices seen in the first $i-1$ observations. Since this is an exchangeable distribution, given $n$ observations $ X = \left[x_1 , \ldots , x_n \right]^{T}$, the conditional distribution over the $i$th feature indicator takes the form
\begin{equation}
    Pr(z_i=k|X, Z_{-i}) \propto \begin{cases} n_{-ik} \cdot f(x_i|\theta_k) & k\in K^+\\
    \alpha \int_{\Theta} f(x_i|\theta)d\theta & k\not \in K^+\end{cases}\label{eqn:crp_conditional}
\end{equation}
where $Z_{-i}$ is the set of all feature indicators excluding the $i$th; $n_{-ik} = \sum_{j\neq i}\mathbb{I}(z_j=k)$, and $Z^{+}$ is the set of unique values in $Z_{-i}$.

The infinite number of features assumed \textit{a priori} by the Dirichlet process is attractive in a clustering/mixture modeling scenario because it allows us to avoid pre-specifying a fixed number of features. While the posterior distribution is also an infinite-dimensional measure, only a finite (but random) number of the infinitely many atoms will be associated with data. In other words, we can think of a DPMM as providing an implicit posterior distribution over the number of features used to represent our data, while allowing for previously unseen features that represent future data points\footnote{For a more substantial introduction to Bayesian non-parametrics, we direct the reader to \cite{Ghosh:Ramamoorth:2003,Hjort:Holmes:Muller:2010,Muller:Quintana:Jara:Hanson:2015} and a recent survey on Bayesian non-parametrics \citep{Xuan:Lu:Zhang:2019}.}.

\subsection{Inference in the Dirichlet process}
One reason for the popularity of the Dirichlet process is that its finite-dimensional Dirichlet marginals lead to conjugacy to an infinite-dimensional analogue of the multinomial distribution. This allows us to construct relatively straightforward samplers (see \citealp{Neal:1998} for a number of different MCMC sampling techniques for the DPMM). As discussed by \cite{Papaspiliopoulos:Roberts:2008}, these algorithms can be split into uncollapsed and collapsed samplers. 

\subsubsection{Uncollapsed Sampler}
Uncollapsed, or batch, samplers explicitly instantiate the mixing measure $ \pi $. Since $\pi$ is infinite-dimensional, we typically work with a finite-dimensional truncation \citep{Ishwaran:Zarepour:2002}. However, any fixed finite truncation is only an approximation, and will not allow us to introduce additional features beyond the pre-specified truncation level. Instead, we can sample from the exact posterior by sampling a random truncation level through a slice sampler \citep{Damien:Wakefield:Walker:1999,Walker:2007}.

\subsubsection{Collapsed Sampler}
Collapsed, or sequential, samplers integrate out the mixing parameter $ \pi $, using the exchangeable representation in Equation~\ref{eqn:crp_conditional}. This typically allows for better mixing of the MCMC chain \citep{Liu:1994, Ishwaran:James:2001, Papaspiliopoulos:Roberts:2008}. However, one challenge when working with the collapsed representation is calculating the likelihood under a currently uninstantiated feature, $\int_\Theta f(x_i|\theta) d\theta$. When $H$ is not a conjugate prior for our likelihood, a popular choice is the ``Algorithm 8'' proposed by \citet{Neal:1998}. In this algorithm, summarized in Algorithm~\ref{alg:alg_8}, rather than sample a previously unseen feature with probability proportional to $\alpha$, we first sample $m$ new feature parameters from $H$, and select each of these potential features with probability $\alpha/m$.


\subsubsection{Distributed Inference}
In large scale modeling scenarios, it is typically the case that the data set is partitioned over several different computers. Moreover, even if the data set could be moved to one computer, it is more efficient in these scenarios to distribute the data and parallelize the computation. However, MCMC algorithms are generally not trivially parallelizable without introducing excessive variance in the posterior \citep{Williamson:Dubey:Xing:2013}. 

Uncollapsed algorithms that instantiate $\pi$ are inherently parallelizable, since the feature allocation probabilities, $P(z_i=k|-) \propto \pi_k f(x_i|\theta_{k})$ are conditionally independent given the mixing proportions $\boldsymbol{\pi}$ and the feature parameters $\{\theta_k\}$. 
For example, \cite{Ge:Chen:Wan:Ghahramani:2015} develop an an efficient, parallel map-reduce procedure for slice sampling the feature allocations.

Collapsed samplers, however, are more challenging to parallelize, since evaluating the marginalized full conditional in Equation~\ref{eqn:crp_conditional} requires excessive and costly communication between processors. Instead of evaluating this marginalized distribution exactly, some distributed collapsed samplers approximate the feature allocation probability. \cite{Smyth:Welling:Asuncion:2009} propose an asynchronous method for the hierarchical Dirichlet process \citep{Teh:2004} which distributes the data across $ P $ processors and performs Gibbs sampling based on approximate marginalized distribution, $ Pr(z_i| Z_{-i})$. \cite{Broderick:Kulis:Jordan:2013} form a parallelizable approximate algorithm for collapsed sampling using a MAP estimate as a substitute for the marginalized full conditional.

As an alternative to using a fully collapsed or fully uncollapsed distributed sampler, we may instead partially collapse out a portion of the mixing measure. This type of partially collapsed method retains the inherently parallelizable structure of uncollapsed samplers while utilizing the mixing speed of collapsed samplers. \cite{Williamson:Dubey:Xing:2013} propose an exact sampler for the DPMM, but this algorithm requires that each observation associated with a particular feature must exist on the same processor. \cite{Zhang:2016} developed an exact sampler for completely random measures (CRM) which exploits the conditional independencies of the features by partitioning the random measure into a finite instantiated partition and an infinite uninstantiated partition. The instantiated portion runs the inherently parallel sampler with the mixing parameter, $ \pi $, instantiated and at random one processor is selected to sample and propose new features based on the predictive distribution of the cluster assignments.






Many of these distributed methods perform well in low-dimensional spaces. However, they tend to struggle as the number of dimensions increases. This is because they all primarily address the issue of sampling the full conditional feature allocations, $Pr(Z|-)$ for $ Z = (z_1, \ldots , z_n)  $, and assume the existence of an appropriate method for sampling parameters $\theta$. Efficient exploration of the feature parameter space requires a method for sampling from the base measure $H$ in a manner that encourages samples with high posterior probability. If $H$ is conjugate to our likelihood family, we can integrate out the feature parameter for a new feature, which will typically allow us to explore the space well. However, if $H$ is not conjugate to the likelihood, such a collapsed representation for new features is not typically analytically available. An alternative is to propose features from an \textit{uncollapsed} representation---meaning, drawing new feature parameters $\theta^*$ from the base measure, $ H $, and assigning observations to clusters with likelihood $ f(X|\theta^*) $. While this avoids the need to integrate out $\theta$, in a high-dimensional space this can lead to very slow mixing, since randomly sampled features are unlikely to be close to the data. 



\subsubsection{Split-merge moves for Bayesian non-parametric models}
One way to improve mixing in MCMC-based algorithms is to propose split-merge moves to accelerate feature learning in mixture models \citep{green2009reversible}. The split-merge proposals represent bigger transitions in the posterior as opposed to the local move in typical Gibbs samplers which may get stuck into local modes. \cite{Jain:Neal:2004} introduced a split-merge method  where the transition probability for an Metropolis-Hasting split-merge proposal simplifies due to conjugacy between the likelihood and prior. \cite{Jain:Neal:2007,Dahl:2005} developed proposals for splitting and merging clusters in Dirichlet process mixture models in MCMC that is applicable to non-conjugate likelihoods. \cite{Chang:Fisher:2013} developed a split-merge approach that is amenable for parallelization.  \cite{Ueda:Ghahramani:2002} proposed a method for split-merges of features in a variational setting for finite mixture models while \cite{Hughes:Sudderth:2013} developed a split-merge proposal for stochastic variational inference applied to DPMMs with conjugate likelihoods. Beyond the Dirichlet process, \cite{Fox:Hughes:Sudderth:Jordan:2014} developed a split-merge algorithm for the beta process hidden Markov model, a model based on the beta-Bernoulli process. However, implementing such samplers in practice is generally difficult as they require non-trivial derivations for different BNP priors and different likelihoods.


\section{Method}\label{sec:method}

Our novel sampling method combines distributed parallel sampling for the Dirichlet process with an improved proposal method for new features.  Essentially, we propose new features in parallel, centered at observations that are poorly modeled by their assigned features. We begin by proposing an asymptotically exact, non-distributed sampler in Section~\ref{sec:exact_method}. While this method is not a practical scalable algorithm, it provides inspiration for a two-stage inference method described in Section~\ref{sec:approx_method}. The first stage of this algorithm is an acceleration stage, where features are proposed in parallel to ensure a high-quality pool of potential features. This stage is an approximation to the exact sampler in Section~\ref{sec:exact_method}. To correct for the approximations introduced in this first stage, the method proceeds with a distributed MCMC algorithm that guarantees asymptotic convergence.

\subsection{An exact, serial sampler with data-driven proposals}\label{sec:exact_method}
This method of proposing new features---by sampling $m$ features from the base distribution $H$---essentially means we must, slowly, explore a high dimensional space in the hope of finding meaningful features. Despite being guaranteed to converge asymptotically, we would not expect this inference strategy to converge in a reasonable amount of time, particularly as the dimensionality of $\Theta$ grows. This is because only a small portion of the parameter space has a non-negligible amount of posterior mass to adequately model a high dimensional data set.


%

We follow Neal in sampling the $|K^+|$ instantiated features, but diverge in how we sample the new parameters associated with the ``empty'' features associated with $K^{-}$. Recall that Neal's Algorithm 8 (pseudo-code written in Algorithm~\ref{alg:alg_8}) is a collapsed sampler where, at each iteration, an observation's parameter is selected from the union of the $K^+$ parameters associated with other data points, and a finite set of $K^-$ ``unobserved'', but instantiated, parameters sampled from the base measure $H$. Let $ K^{+} $ represent $\left\{k : \sum_{i=1}^{N}\mathbf{I}(z_{i}=k) > 0  \right\}$, the index set of allocated features at the current MCMC iterations, and let $K^{-}$ indicate the indices of a finite set of parameters that are not currently associated with any data, but which are instantiated by the sampler. At the beginning of each iteration, these ``unoccupied'' parameters are replaced with $m$ new parameters, sampled i.i.d.\ from $H$, where $m$ is a user-defined constant. New values for ``occupied'' parameters $\{\theta_k: k\in K^+\}$ are sampled from their appropriate conditional distributions. Next, we iterate through the data, sampling feature  assignment indicatorss from $\{z_k: k\in K^+ \cup K^-\}$. As we sample these feature assignment indicatorss, the numbers $|K^+|$ and $|K^-|$ of ``occupied'' and ``unoccupied'' features will change,  as observations join and leave features. 
Instead of sampling empty features from the base measure $H$, we instead propose a new set of feature locations $\{\theta^*_k: k\in K^-\}$ according to an empirical feature proposal algorithm, as seen in Algorithm~\ref{alg:acc_exat} below, and accept or reject these features according to a Metropolis-Hastings acceptance ratio.
%

\begin{algorithm}[h!]
	\label{alg:alg_8}
	\caption{Algorithm 8 from \cite{Neal:1998}.}
	\For{$i=1, \ldots , N$}{
		\For{$k \in K^{-}$}{
			Sample $\theta_{k} \sim H$
		}
		Sample \begin{equation*}
		Pr(z_i=k | -) \propto \left\{ \begin{array}{cc}
		n_{-ik} \cdot f(x_i | \theta_{k}), & k \in K^{+}\\
		(\alpha / m) \cdot f(x_i | \theta_{k}), & k \in K^{-}
		\end{array} \right.
		\end{equation*}
	}
	\For{$k \in K^{+}$}{
		Sample $\theta_{k} \sim Pr(\theta_k | - )$
	}
\end{algorithm}

We begin by defining the start point of our Metropolis step. Following a slight modification of Neal's Algorithm 8, we ensure that $|K^-|=m$, by either randomly deleting unoccupied locations if the current number is greater than $m$, or by sampling new locations from $H$ if the current number is less than $m$; we could alternatively sample a new set of $m$ locations from $H$. This step simplifies the final Metropolis Hastings ratio, by ensuring the proposal does not change the dimensionality.

Then, for each $k\in K^-$ in turn, we propose a candidate location $\theta^*$, sampled from a mixture distribution,
\begin{equation}
Q = (1-\rho) H + \rho \sum_{i=1}^{n} \delta_{\tilde{\theta}_{x_i}}  \frac{ 1/ f  \left( x_i | \theta_{z_i} \right) }{\sum_{i=1}^{n} (1/f \left( x_i | \theta_{z_i} \right)) }.\label{eqn:empirical}
\end{equation}
where $\tilde{\theta}_{x_i}$ is the the data point $x_i$ projected to the parameter space for the given likelihood (see later for a more detailed discussion). This distribution is designed to concentrate probability mass in areas that are near the data, but are not well-represented by the parameters in $K^+$. With probability $1-\rho$, we sample from the prior, and with probability $\rho$, we sample from an empirical distribution whose support corresponds to that of the data. The probabilities assigned with each of the data locations is inversely proportional to the data points likelihood given its current feature assignment, $f(x_i|\theta_{z_i})$.  Any alternative distribution that concentrates mass near poorly explained observations could be substituted here; we choose $ 1/f(\cdot) $ since it is an  easily interpretable way to concentrate probability near observations that are poorly explained by their current feature assignment

To ensure validity of the sampler, we accept the feature proposal $\theta^*$ using the Metropolis-Hastings acceptance probability
\begin{equation}
    \min\left\{1, \frac{H(\theta^*_k)Q(\theta_k)}{H(\theta_k)Q(\theta^*_k)}\right\}\label{eqn:MH}
\end{equation}
This mixture distribution and Metropolis-Hastings correction ensures that our sampler is still irreducible, meaning there is non-zero probability of $\theta_k$ transitioning to any other state on the parameter space's support. If $\rho=0$, we are proposing feature locations directly from $H$, meaning that the Metropolis Hastings acceptance probability simplifies to one and we automatically accept any draw from the prior, as we would typically do in Algorithm 8. The Metropolis-Hastings sampling step for uninstantiated features is described in Algorithm~\ref{alg:acc_exat}.

Practically, this method displays a flavor similar to split moves found in split-merge algorithms \citep{Chang:Fisher:2013,Jain:Neal:2007}, where we hope that data points who are currently poorly represented by their feature assignment can be split off into a new feature. In split-merge algorithms, this occurs by explicitly assigning data to a new feature, and calculating the corresponding acceptance probabilities, which can be challenging to derive and implement. By contrast, we are proposing replacing unoccupied features, meaning the acceptance ratio is much more straightforward. Moreover, since the relevant features are unoccupied, our method can easily be used with non-conjugate likelihoods. 


The performance of feature proposal algorithm depends on having a way of selecting parameter values $\tilde{\theta}_{x_i}$ that in some way ``resembles'' the data point $x_i$. Appropriate choices here are easy to specify when the maximum likelihood or maximum a posteriori value for $\theta$ is available given a single observation $x_i$. This is often the case for location parameters, such as the mean of a Gaussian, the proportions of a multinomial, or the parameter of a Poisson. Where maximum likelihood estimators are not available, such as the gamma and beta distributions, we may be able to estimate a single parameter using techniques such as the method-of-moments. 

However, it is still not clear how to learn higher order moment parameters. We argue that accelerating the mixing of such samplers depends more on accelerated learning of the proper locations than the scale parameters. But in such cases, we can propose new features centered at the given observation and the covariance of its previous assignment, rescaled to be more tightly centered at its mean (in the Normal-Inverse Wishart case, for example). The purpose of this proposal is so that data assigned to this feature are more similar to the new proposed feature, and we learn the covariance matrix by sampling posterior updates for the features in an exact fashion (like Inverse-Wishart conjugate updates) to obtain the correct posterior covariance.


\begin{algorithm}
	\label{alg:acc_exat}
	\caption{Empirical Feature Proposal Algorithm}
	\For{$k \in  K^{-} $ }{
			Sample $ u \sim \mbox{Uniform}(0,1) $\\
			\If{$ u < \rho $}{
				Set proposal distribution $Q$ to an empirical distribution of the data:
				\begin{equation*}
				Q := \sum_{i=1}^{n} \delta_{\tilde{\theta}_{x_i}} \frac{1/ f  \left( x_i | \theta_{z_i} \right) }{\sum_{i=1}^{n} (1/f \left( x_i | \theta_{z_i} \right)) },  	
				\end{equation*}

			}
			\Else{
				Set $ Q := H $ (the base measure)
			}
			Draw $ \theta^{\ast} \sim Q  $ and accept new state for $ \theta_k $ with probability $ \min\left\{ 1, \beta \right\} $ where
			\begin{equation*}
			\beta = \frac{ H(\theta^{\ast})Q(\theta_k)}{H(\theta_k)Q(\theta^{\ast})}
			\end{equation*}
	}
\end{algorithm}

\subsection{An approximate distributed inference algorithm}\label{sec:acceleration}
The exact sampler described above requires access to all of the data on a single processor--both in sampling from the empirical distribution, $\sum_{i=1}^{n} \delta_{\tilde{\theta}_{x_i}} \frac{1/ f  \left( x_i | \theta_{z_i} \right) }{\sum_{i=1}^{n} (1/f \left( x_i | \theta_{z_i} \right)) }$, and sampling the collapsed feature assignments $ Pr(Z|-) $. This makes the algorithm inherently unparallelizable. Worse yet, as the dimensionality of the data grows the probability of accepting a state sampled from the data goes to zero. To overcome these issues, next we propose an easily parallelizable approximation of our accelerated algorithm.

Our approximation consists of two components. The first is an approximation of the collapsed sampling algorithm where we replace the global summary statistics necessary for the sampler with summary statistics that are local to the processor. Here, the global summary statistics used to calculate the full conditional, $n_{-ik}$ refers to the number of observations allocated to feature $k$ except for observation $i$. For newly introduced features in this accelerated stage, we assume we observe this feature only on its given processor. Thus, we only use the local summary statistics to update the new features. We do not need to approximate the global summary statistics for the instantiated feature parameters here as we send the local statistics to a central processor and exactly sample the posterior feature update and transmit the new state to all the other processors. 

In a single processor setting, we will always have the exact value of $n_{-ik} $. However, in the parallel setting we no longer have the precise feature counts, which forces inter-processor communication for every state transition of $ z_i $. Instead, we approximate the global summary statistics as $ P \cdot n_{-ikp} $, which is the local count of feature $ k $ on processor $ p $ excluding observation $ i $ when running the algorithm in parallel, multiplied by the number of processors used. In other words, given the local feature allocations, and the fact that observations are randomly assigned to processors we approximate $n_{-ik} $ with $E[n_{-ik}|n_{-ikp}]$.

We choose to use this approximation because collapsed samplers, though inherently unparallelizable, generally exhibit better mixing properties than uncollapsed ones \citep{Liu:1994, Papaspiliopoulos:Roberts:2008}. \cite{Smyth:Welling:Asuncion:2009} introduce this approximation in their distributable sampler and demonstrate that this approximation, though an incorrect sampler, still produces good empirical results.  Explicitly, we accelerate sampling with a modified version of Algorithm~8 \citep{Neal:1998} which introduces $ m $ auxiliary features $\theta_k $ for $ k=1 , \ldots , m $ that represent finite realizations of the features. Algorithm~8 assigns feature assignments according to the following probability:
\begin{equation}
Pr(z_i=k | -) \propto \left\{ \begin{array}{cc}
	P \cdot n_{-ik} \cdot f(x_i | \theta_{k}), & k \in K^{+}\\
	(\alpha / m) \cdot f(x_i | \theta_{k}), & k \in K^{-}
\end{array} \right.
\end{equation}\label{eqn:Qdist}
Then, draw new realizations for $ \theta_k, k \in K^{-} $ from $ H $ and update posterior values for $ \theta_k , k \in K^{+} $.



The second approximation is that, rather than use the Metropolis-Hastings acceptance ratio in Equation~\ref{eqn:MH} to accept proposed unoccupied features, we automatically accept all proposed features from the empirical distribution in Equation~\ref{eqn:empirical}. Using the exact proposal algorithm in Alg.~\ref{alg:acc_exat} results in low acceptance probabilities in high dimension. This is because the ratio of proposal densities in the Metropolis-Hastings acceptance probability discourages large moves in the parameters space, and the penalization of large moves becomes even greater in higher dimensions \citep{Au:Beck:2001,Katafygiotis:Zuev:2008}. Therefore, in the exact situation where we need this feature proposal algorithm is the one where we end up sampling from the prior if we use the exact method.  In practice, we are instead directly, empirically sampling $m$ new features at each iteration, according to Equation~\ref{eqn:Qdist}, in order to minimize the amount of time wasted exploring portions of the feature space that lack useful information about the data. This approximation sacrifices the theoretical correctness guaranteed by the Metropolis-Hastings step, in order to more quickly explore the state space during the accelerated stage of the sampler. Despite that this initial stage is only an approximation, we can see in the experimental results that this accelerated method will invariably reach a local mode faster in contrast to exact DP samplers that sample feature locations from $ H $ or assign feature allocations based on $\int_\Theta f(x_i|\theta) d\theta$.

We assume that new features accepted on different processors are different from each other, thus we do not consider the problem of feature alignment. Furthermore, by dividing computation across multiple processors we can propose $ P $ times more features to explore possible new features that will persist after acceleration. After $ L $ subiterations of our sampler, we trigger a global synchronization step where each processor sends updated features, feature counts and feature summary statistics to instantiate new features on all processors and to update posterior values for global parameters.

\subsection{A two-stage, asymptotically exact parallel inference procedure}\label{sec:approx_method}
The method described in Section~\ref{sec:acceleration} is obviously not a correct MCMC sampler for a DPMM since it does not apply appropriate Metropolis-Hastings corrections. However, we can use it as an acceleration stage in a two-stage algorithm. Moreover, in the parallel setting, the reliance of local counts is an approximation of the total counts of observations assigned to feature $ k $. Any asymptotically correct MCMC sampler for the DPMM will eventually converge to the true posterior, regardless of its starting point---but the time required to achieve arbitrary closeness to this posterior will be strongly dependent on that start point. We begin our inference procedure by running the approximate accelerated sampler for some pre-determined number of iterations, and then switch to an exact distributable inference algorithm. This corresponds to using the empirical feature proposal distribution from Algorithm~\ref{alg:acc_exat} with $ \rho=1 $ for the accelerated stage and $ \rho=0 $ for the exact stage. 

We choose to use the distributed sampler of \citealp{Zhang:2016} in the exact stage of the sampler. The algorithm by \cite{Zhang:2016} instantiates the mixing measure, $ \pi_k $, for the features that have data assigned to them. Otherwise the algorithm integrates out the infinite dimensional tail of unoccupied features. This approach is applicable to any completely random measure (CRM), where we may partition the latent measure, and for transformations of CRMs, in the case of the Dirichlet process, where we can decompose the DP into a $ B\sim\mbox{Beta}(N,\alpha) $ mixture of the two parts \citep{Ferguson:1973,Ghosh:Ramamoorth:2003}. Moreover, we can use any choice of likelihood and base measure using this algorithm, unlike other MCMC samplers which may depend on using conjugate priors.

The major benefit of our accelerated sampler is that we have an efficient sampler to encourage fast mixing of the MCMC sampler without the need to integrate the latent features out of the likelihood. Thus, we can now use a variety of priors for features without encountering the problem of proposing features from the prior in high dimensional space. Our method in the DPMM case is suitably general for a wide class of data modeling scenarios. Although the most common type of mixture is the Gaussian mixture model, we do not place any assumption on the form of likelihood for $ f(X|\theta) $, and we will see an example of our method applied to count data to demonstrate the flexibility of this method. 

Algorithm~\ref{alg:accelerated} provides the exact details for the inference procedure of our accelerated sampler for one iteration. We denote $ K^{+} $ to mean the set of features instantiated on all processors, $ K^{\ast} $ to mean the set of newly introduced features not yet instantiated on all processors, and $ K^{-} $ as the set to represent the unoccupied features. In the accelerated stage, each processor is allowed to accept new features--hence the subscript $ p $ in $ K^{\ast}_p $ and $ K^{-}_{p} $. Once the accelerated stage ends, only one processor $ p^{\ast} $, drawn uniformly, may sample new features while all other processor may only sample instantiated features. Furthermore, we assign feature indicators for the features in $ K^{+} $ according to an uncollapsed sampler in order to maintain theoretical correctness of the algorithm. Lastly, if the current iteration triggers a synchronization step, then all processors send summary statistics and new features to the master processor. The master processor samples new states for the global parameters, $ (\pi, \alpha, \theta) $. We sample the instantiated $ \pi_k $, from its conjugate update, and we sample the posterior of $ \alpha $ from the method described in \cite{West:1992}.  For $ \theta $, if the choice of base measure $ H $ is conjugate to the likelihood of the model then we can sample accordingly from its conjugate update, otherwise we must update $ \theta $ according to a non-conjugate method (ex: Metropolis-Hastings, Hamiltonian Monte Carlo). After sampling the global parameters, we send the updated parameters and newly introduced features to all the processors.

\begin{algorithm}
	\label{alg:accelerated}
	\caption{Accelerated DP Inference Algorithm}
	\If{accelerated}{
		\For{$ p=1 , \ldots , P $ in parallel}{
			\For{$ i = 1, \ldots , n_p $}{Sample \begin{equation*}
			Pr(z_i=k | -) \propto \left\{ \begin{array}{cc}
			\frac{P \cdot n_{-ikp}}{N + \alpha -1} \cdot f(x_i | \theta_{k}), & k \in K^{+}\\
			\frac{n_{-ik}}{N + \alpha -1} \cdot f(x_i | \theta_{k}), & k \in K^{\ast}_{p}\\
			\frac{\alpha / m }{N + \alpha -1} \cdot f(x_i | \theta_{k}), & k \in K^{-}_{p}
			\end{array} \right.
			\end{equation*}
			}
			\For{$ k \in K^{-}_{p} $}{
			$\theta_{k} \sim \sum_{i=1}^{n_p} \delta_{\tilde{\theta}_{x_i}}\frac{ 1/ f  \left( x_i | \theta_{z_i} \right) }{\sum_{i=1}^{n} (1/f \left( x_i | \theta_{z_i} \right)) }$
			
			}		
		}
	}
	\Else{
		\For{$ p=1 , \ldots , P $ in parallel}{
            Sample according to \cite{Zhang:2016}.
		}		
	}
	\If{synchronization step }{
		On master processor, gather summary statistics and new instantiated features from all processors.\\
		$ K^{+} : = K^{+} \cup_{p=1}^{P}  K^{\star}_{p} $\\
		Sample instantiated mixture weights, $\{ \pi_k : k \in Z^{+}\} \sim \mbox{Dirichlet}( \{ n_k : k \in Z^{+} \} )$\\
		Sample $\alpha\sim\mbox{Gamma}(\alpha_0+ |K^{+}| - 1, \beta_0 + g + \log N )$, where $ g $ is Euler's constant.\\
		Sample $B\sim \mbox{Beta}(N,\alpha)$.\\
		Sample $\theta \sim Pr(\theta|-) $.\\
		Sample $ p^{\ast} \sim \mbox{Uniform}(1, \ldots , P) $.\\
		Transmit $ \left\{\pi, \theta, \alpha , p^{\ast} , K^{+} \right\} $ to new processors.
	}
	
%
	
\end{algorithm}

\section{Experimental Results}\label{sec:experiments}
\subsection{Synthetic example}\label{subsec:synthetic}
\begin{figure*}[ht!]
	\centering
	\includegraphics[width=1.0\linewidth]{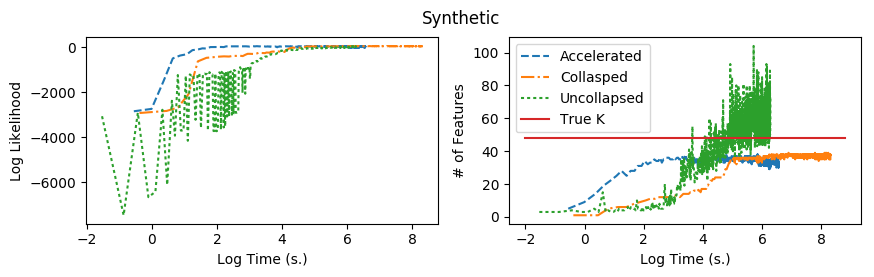}
	\caption{Test set predictive log likelihood vs. log time (seconds) and number of features vs. log time for the multinomial-Dirichlet model.}
	\label{fig:synthetic}
\end{figure*}

As a basic evaluation of our accelerated sampler, we first generated synthetic 10 dimensional data with 1,000 training observations and 100 test observations and a total count of $N_i = 100$ for each observation, $i$, from a Dirichlet mixture of multinomials:
\begin{align}
\begin{split}
f(X_i | \theta_{z_i}, z_i) &\sim \mbox{Multinomial}(N_i, \theta_{z_i}),\\
H &= \mbox{Dirichlet}(\gamma , \ldots , \gamma)
\end{split}\label{eqn:dpmm_mult}
\end{align}
and applied our accelerated sampler from Algorithm~\ref{alg:accelerated} to this data set. Here, we fixed $ \gamma $ to $1$. Even in this simple example, we can see that our accelerated sampler performs favorably in comparison to the collapsed sampler (Neal's Algorithm 8) and uncollapsed sampler (\citeauthor{Ge:Chen:Wan:Ghahramani:2015}'s distributable slice sampler) with regards to convergence time. Though our method may seem like it would trivially grow to place each observation in its own cluster, using the exact stage of the sampler means that our method does not overfit the number of features. We do not expect our sampler to recover the generating number of features, since the DPMM is not a consistent estimator for the number of features in a finite mixture model \citep{Miller:Harrison:2013}. However, we do expect it to asymptotically sample from the true posterior. Figure~\ref{fig:synthetic} suggests that our method converges to a similar stationary distribution as the other exact inference methods.

In a small, low-dimension setting we can easily run the MCMC sampler to convergence. However, for a big data scenario with high-dimensional data, we cannot realistically run the sampler until convergence especially with consideration to fixed computational budgets. In the next section, we will motivate the necessity for our accelerated MCMC method in this big data scenario by examining three image data sets that fail to mix or mix slowly with traditional sampling techniques.

\subsection{Image data sets}\label{subsec:image}
We will apply our accelerated inference technique on three large dimensional data sets:\footnote{Code is available at \texttt{https://github.com/michaelzhang01/acceleratedDP}}
\begin{itemize}
    \item The $ 28 \times 28 $ dimensional MNIST handwritten digit dataset \citep{LeCun:Cortes:1998}, consisting of a training set of 60,000 images and test set of 10,000  images.
    \item The $ 168 \times 192 $ dimensional Cropped Extended Yale Face Dataset B \citep{Lee:Ho:Kriegman:2005}, divided into 1,818  training images and 606 test set images.
    \item The CIFAR-10 \citep{Krizhevsky:2009} image data set converted to greyscale with 50,000 training images and 10,000 test images.
\end{itemize}

For these image datasets, we model the data with a multinomial likelihood and a Dirichlet base measure. Each image is represented as a $D$-dimensional vectors $X_i$, where each entry $x_{id}$ is an integer from 0 to 255, indicating the pixel intensity of image $n$'s $d$-th pixel and each image as having a total pixel sum as $N_i = \sum_d x_{id}$ which we assume to be known. We choose to use this likelihood because we can take advantage of multinomial-Dirichlet conjugacy and easily compare with other methods that depend on conjugate priors. However, this model choice has numerous obvious flaws--there is no spatial dependence among pixels, the likelihood does not strictly limit the value of the pixels to be less than $255$, and the likelihood is sensitive to rotations and scaling of the data (this drawback becomes particularly noticeable in the MNIST example where the learned clusters are rotated and rescaled variants of the digits). 

We compare our results to the mean-field variational inference algorithm for learning DP mixtures from \cite{Blei:Jordan:2006}. Variational methods are typically faster than MCMC algorithms, but lack the theoretical guarantees. In each experiment we initialized all observations to the same cluster and distributed the data randomly across $ 10 $ processors for the distributable algorithms (our method and the uncollapsed sampler). We ran the sampler for either $ 1000 $ iterations or two days (whichever arrived first) with a global synchronization step every $ 10 $ iterations (synchronization steps are used for our method and the uncollapsed sampler only) and stopped the accelerated sampling after $ 50 $ iterations.

To show that our acceleration method provides benefits beyond a naive initial pre-clustering, we explored several initializations for each algorithm. Rather than initialize to a single cluster, we initialized to either $K=100$ initial clusters, either via random sampling, or by $K$-means clustering (labeled ``Rand. Init.'' and ``KM Init.'' on the figures, respectively). As in the previous section, we set $ \gamma=1 $. We learned $ \alpha $ according to \cite{West:1992} for the MCMC-based methods but fixed $ \alpha=10 $ for the variational experiments. As a baseline comparison, we also run only the hybrid parallel DPMM algorithm from \cite{Zhang:2016}, which constitutes the second exact stage of our sampler, initialized with all the data in a single cluster.

\begin{figure*}[ht!]
	\centering
	\includegraphics[width=1.0\linewidth]{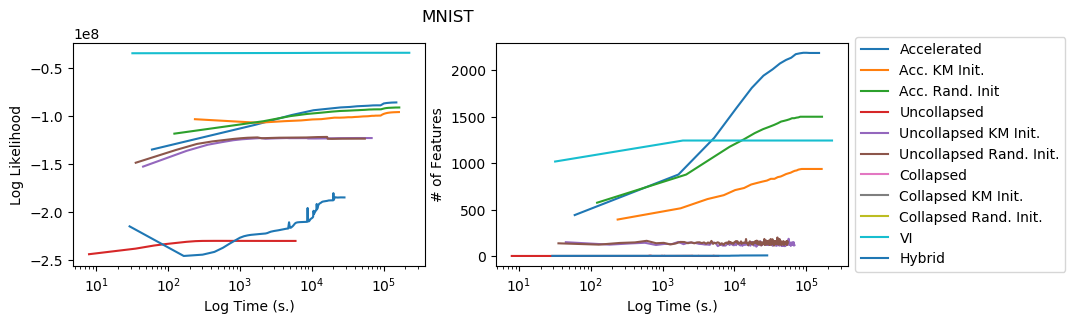}
	\caption{Test set predictive log likelihood vs. log time (seconds) and number of features vs. log time for the multinomial-Dirichlet model.}	\label{fig:MNIST_comparison}
\end{figure*}

\begin{figure*}[ht!]
	\centering
	\includegraphics[width=1.0\linewidth]{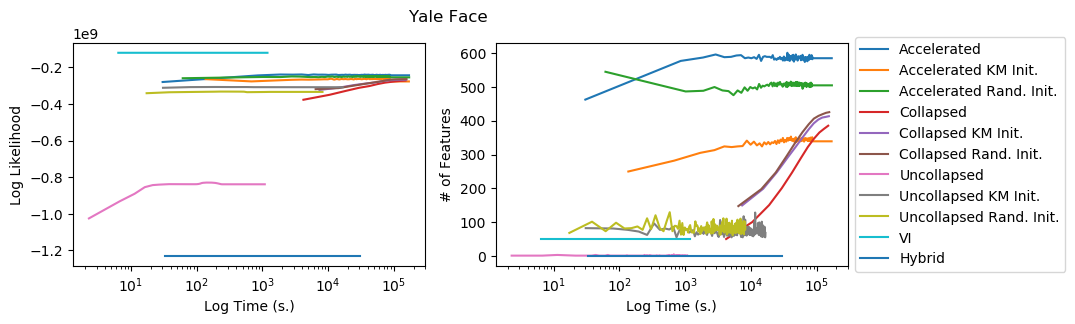}
	\caption{Test set predictive log likelihood vs. log time (seconds) and number of features vs. log time for the multinomial-Dirichlet model.}	\label{fig:yale_comparison}
	
\end{figure*}

\begin{figure*}[ht!]
	\centering
	\includegraphics[width=1.0\linewidth]{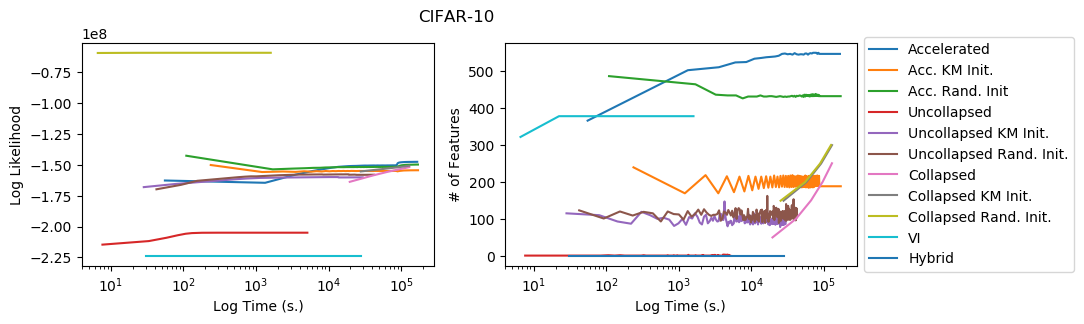}
	\caption{Test set predictive log likelihood vs. log time (seconds) and number of features vs. log time for the multinomial-Dirichlet model.}	\label{fig:cifar_comparison}
	
\end{figure*}

For each of the datasets, we can see that the uncollapsed samplers have difficulty proposing good features in high dimensional space. In terms of predictive log likelihood, we perform favorably for each of the data sets against all the sampling methods except against the variational inference. For the uncollapsed sampler, the results demonstrate that it is difficult for the sampler to propose good locations from the prior in high dimensional space and as a result, very few new features are instantiated relative to the size of the data. While the collapsed sampler can still propose good features, the timing results in Figures~\ref{fig:MNIST_comparison}, \ref{fig:yale_comparison}, \ref{fig:cifar_comparison} show that Neal's Algorithm 8 is far slower than the accelerated algorithm with regards to convergence time, due to the fact that it is not inherently a parallelizable sampling algorithm. In fact, the accelerated sampler never even reaches convergence to a local mode in any of the image data set experiments. Even with the parallel hybrid method from \cite{Zhang:2016}, the sampler has trouble proposing new features, as evidenced by the lack of new features throughout the duration of the sampler, which suggests that having an effective feature proposal mechanism is crucial for exploring the posterior space.

\begin{figure*}[ht!]
	\centering
	\includegraphics[width=.3\linewidth]{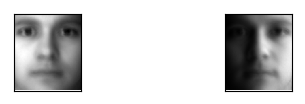}\includegraphics[width=.25\linewidth]{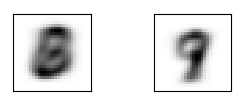}\includegraphics[width=.20\linewidth]{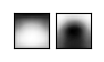}
	\caption{Yale faces features (left), MNIST features (middle) and CIFAR features (right) obtained via uncollapsed sampling, sorted in descending order of popularity for the multinomial-Dirichlet model.}
	\label{fig:yale_mnist_feat_uncollapsed}
\end{figure*}
\begin{figure*}[ht!]
	\centering
	\includegraphics[width=.95\linewidth]{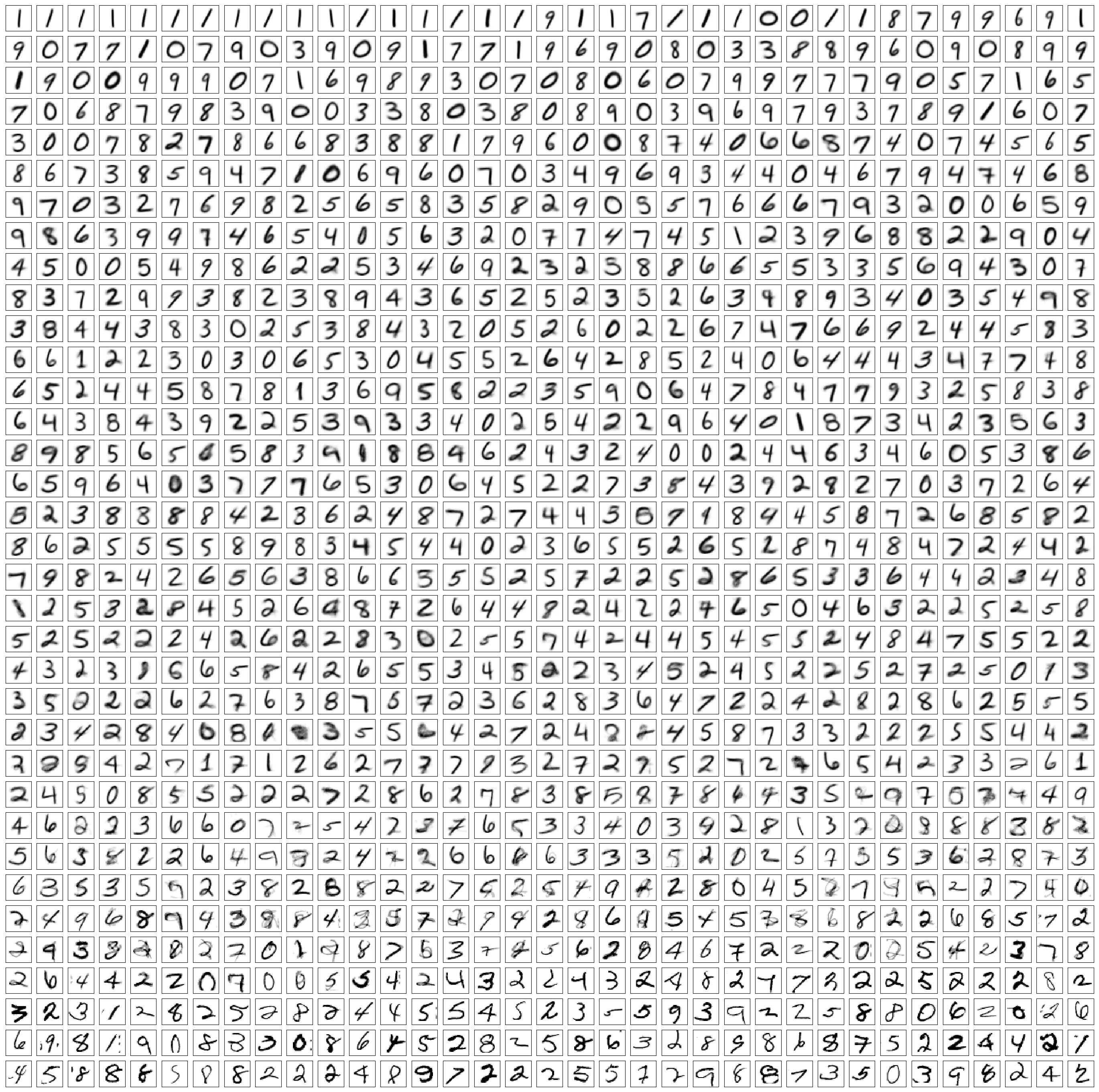}
	\caption{MNIST features obtained via variational inference, sorted in descending order of popularity for the multinomial-Dirichlet model.}
	\label{fig:mnist_vi_feat}
\end{figure*}
\begin{figure*}[ht!]
	\centering
	\includegraphics[width=1.0\linewidth]{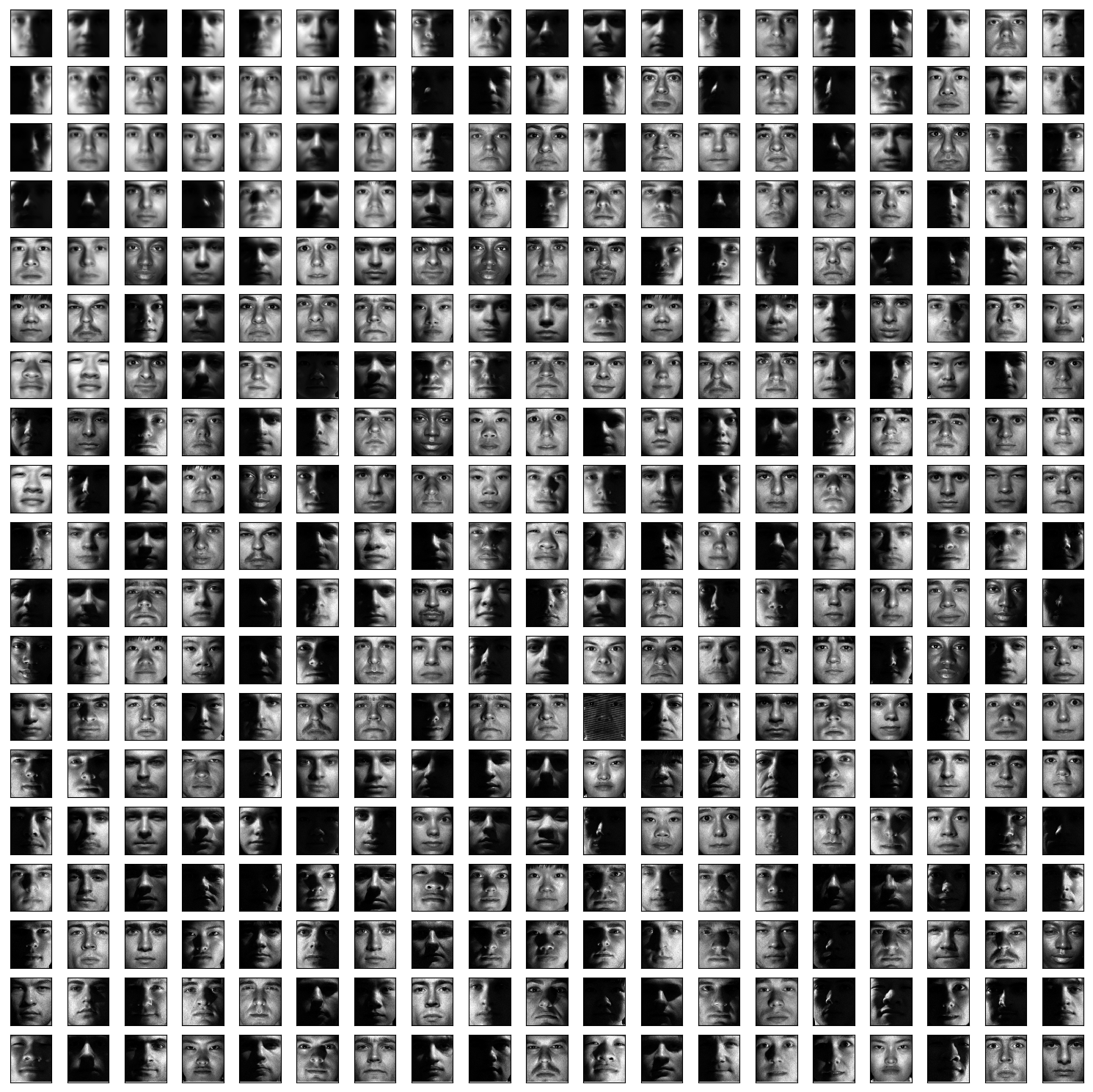}
	\caption{Yale faces features obtained via variational inference, sorted in descending order of popularity for the multinomial-Dirichlet model.}
	\label{fig:yale_vi_feat}
\end{figure*}
\begin{figure*}[ht!]
	\centering
	\includegraphics[width=1.0\linewidth]{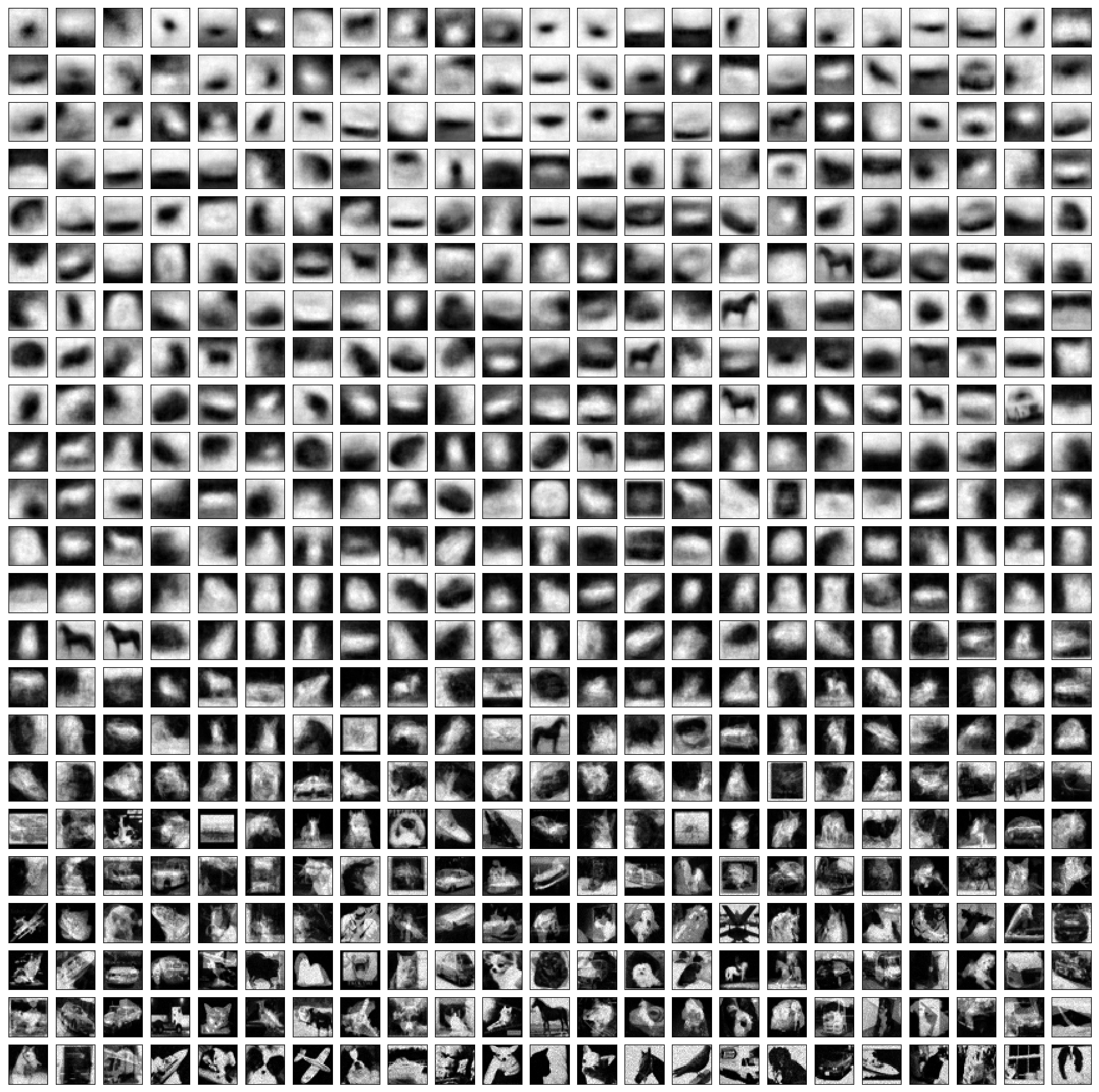}
	\caption{CIFAR features obtained via accelerated sampling, sorted in descending order of popularity for the multinomial-Dirichlet model.}
	\label{fig:cifar_vi_feat}
\end{figure*}
\begin{figure*}[ht!]
	\centering
	\includegraphics[width=.7\linewidth]{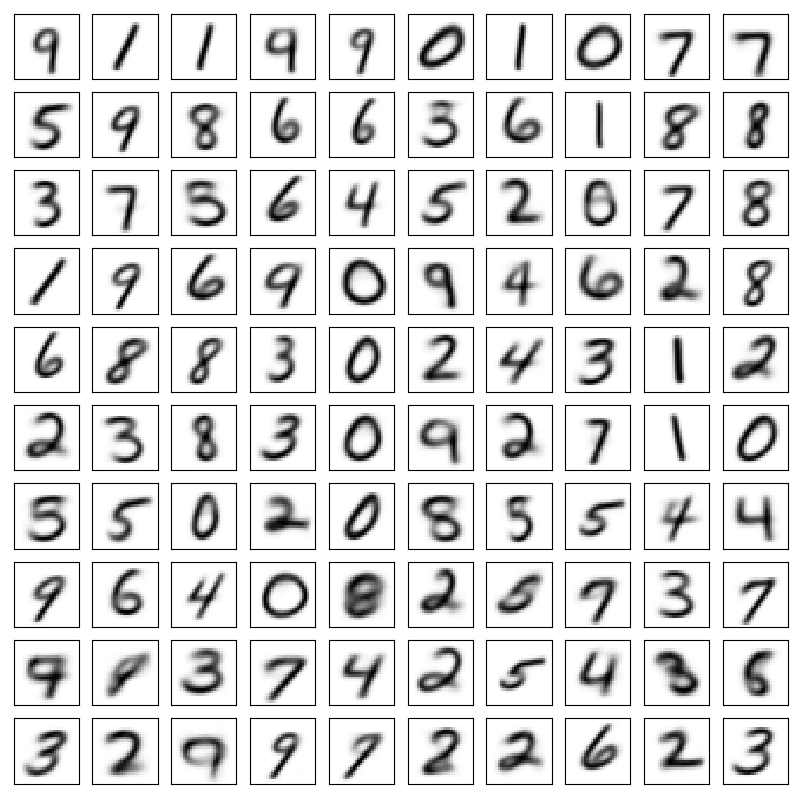}
	\caption{MNIST features obtained via collapsed sampling, sorted in descending order of popularity for the multinomial-Dirichlet model.}
	\label{fig:mnist_feat_collapsed}
\end{figure*}

Moreover, the accelerated features are clearly superior to the ones learned in the uncollapsed algorithm (see Figure~\ref{fig:yale_mnist_feat_uncollapsed}) and are comparable to the ones learned in the variational results (Figures~\ref{fig:mnist_vi_feat}, \ref{fig:yale_vi_feat}, \ref{fig:cifar_vi_feat}). In comparison to the features learned in the collapsed algorithm, our accelerated algorithm has mixed much better than the collapsed algorithm and this is apparent when inspecting some of the collapsed features which tend to be less sharp or contain obviously mis-clustered observations (the fourth feature learned in Figure~\ref{fig:mnist_feat_collapsed} is a composite of the number ``four'' and ``nine''). Although the variational inference results indicate that variational inference is faster and performs better than our method in terms of predictive log likelihood, we would like to emphasize that the variational approach is an approximate method of Bayesian inference that typically would perform faster than sampling based approaches whereas our algorithm is an exact method for inference and performs better than the other MCMC methods examined in this paper and, critically, is parallelizable as opposed to the VI and collapsed algorithms.

Furthermore, the results for random and $ K $-means initialization for the uncollapsed and collapsed samplers demonstrate that we need a smarter way of learning new features beyond just having good initial states. The results show that, under these initializations, we can learn some features of the data but we cannot fare as well as we would with the accelerated sampler. We can also see, by the poor quality of the collapsed and uncollapsed samplers, that our base measure has difficulty accurately representing out high-dimensional data set. This difficulty is one which we raised in the first point of our introduction. Thus, we need our accelerated method to learn features because our method better represents complicated datasets.

\begin{table}[ht!]
	\centering
	\caption{Comparison of final predictive log likelihood results for MNIST, Yale, and CIFAR data with Dirichlet-multinomial model.}
	\begin{tabular}{l|rrr}
		Algorithm & MNIST & Yale & CIFAR-10 \\
		\hline
		Accelerated & -8.61e7 & -2.43e8 & -1.48e8\\ 
		Acc. KM Init & -9.62e7 & -2.77e8 & -1.50e8\\ 
		Acc. Rand Init & -9.14e7 & -2.54e8 & -1.54e8\\
		Uncollapsed & -2.30e8 & -8.39e8 & -2.05e8\\ 
		Unc. KM Init. & -1.23e8 & -3.08e8 & -1.60e8\\ 
		Unc. Rand. Init & -1.24e8 & -3.34e8 & -1.58e8\\
		Collapsed & -1.38e8 & -2.72e8 & -1.52e8\\
		Coll. KM Init. & -1.12e8 & -2.67e8 & -1.51e8\\
		Coll. Rand. Init & -1.29e8 & -2.64e8 & -1.51e8\\
		Variational & -3.44e7 & -1.20e8 & -5.91e7\\
		Hybrid & -1.85e8 & -1.23e9 & -2.23e8\\
		\hline
	\end{tabular}
\end{table}

\begin{figure*}[ht!]
	\includegraphics[width=1.0\linewidth]{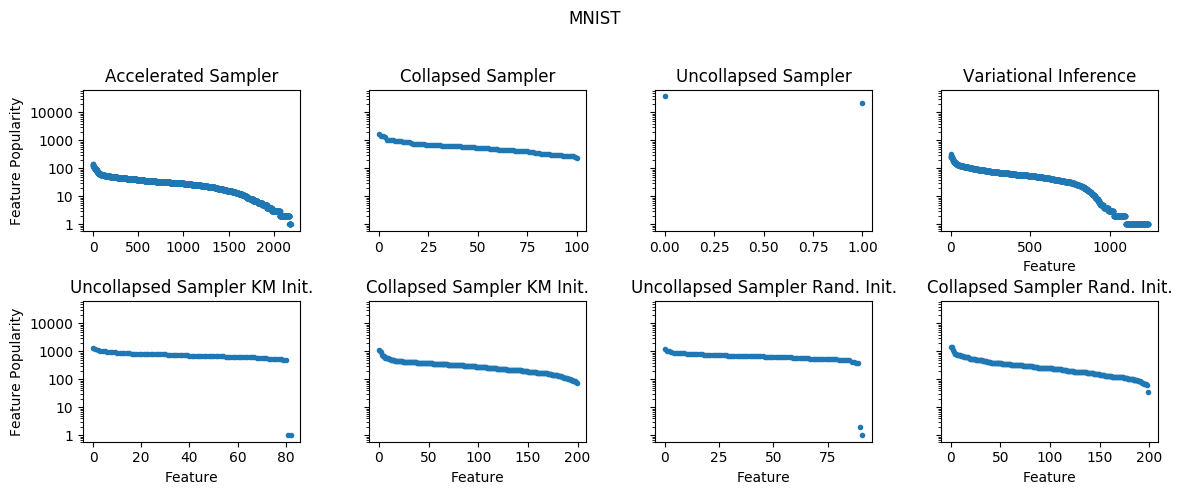}
	\caption{Popularity of each instantiated feature of the MNIST data set for the multinomial-Dirichlet model.}
	\label{fig:MNIST_K}
\end{figure*}

\begin{figure*}[ht!]		
	\includegraphics[width=1.0\linewidth]{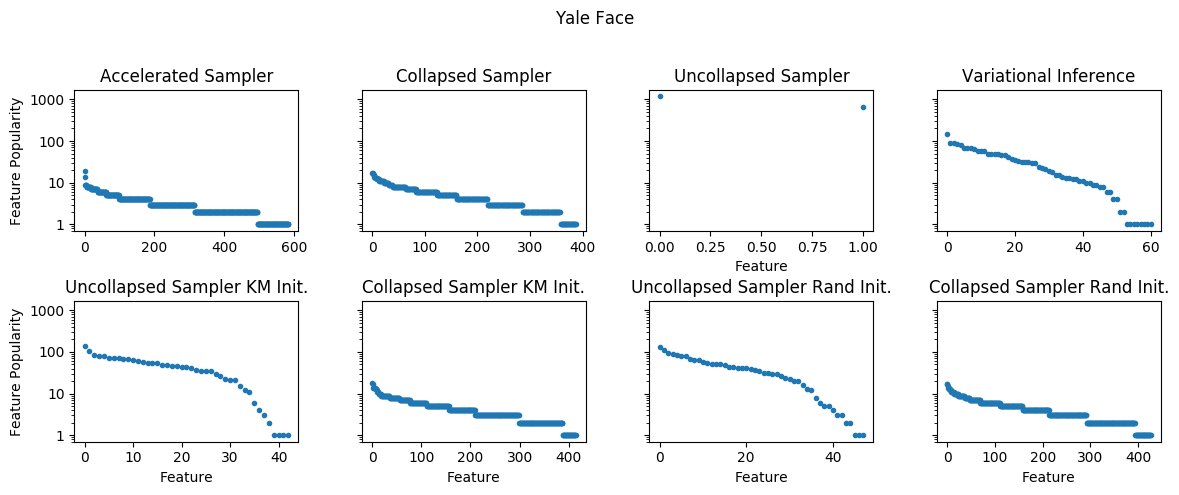}
	\caption{Popularity of each instantiated feature of the Yale data set for the multinomial-Dirichlet model.}
	\label{fig:yale_K}
\end{figure*}

\begin{figure*}[ht!]		
	\includegraphics[width=1.0\linewidth]{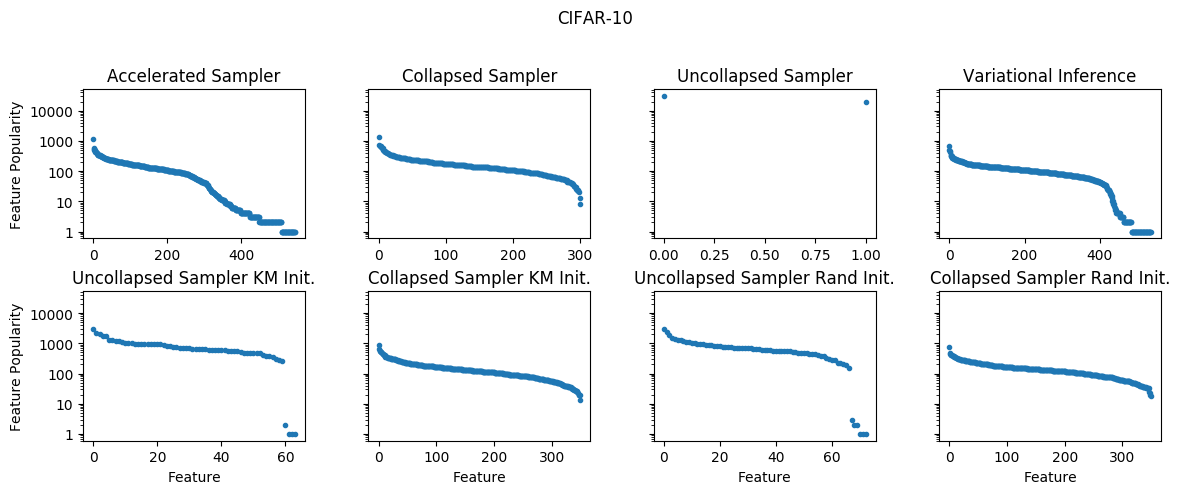}
	\caption{Popularity of each instantiated feature of the CIFAR for the multinomial-Dirichlet model.}
	\label{fig:cifar_K}
\end{figure*}

\subsection{Non-conjugate Experiments}\label{sec:nonconjugate}
We now apply our accelerated sampler for a setting with a non-conjugate prior-likelihood specification. Instead of the Dirichlet prior that we used in the prior experiment, we now assume the base measure is a $ D $-dimensional multivariate normal distribution with mean zero and covariance $ \Sigma $ and the parameter of interest, $ \theta_{k} $, is the multivariate normal from the base measure projected to the $ D-1 $ probability simplex. For memory saving purposes, we assume the covariance parameter $ \Sigma $ is generated from a lower $ r $-rank decomposition, $ \Lambda \Lambda^{T} $, where $ \Lambda $ is a $ D\times r $ matrix generated from independent Gaussians.:
\begin{align}
\begin{split}
f(X_i | \theta_{z_i}, z_i) &\sim \mbox{Multinomial}(\sum_{d}x_{id}, \theta_{z_i}), \\ 
\theta_{k} &=  \frac{\exp\{\mu_{k}\}}{\sum_{d=1}^{D}\exp\{\mu_{k,d}\}}\\
\mu_{k} &\sim H, \; H = \mbox{MVN}(0, \Sigma), \\ 
\Sigma &= \Lambda  \Lambda^{T}, \Lambda_{r} \sim \mbox{MVN}(0, \sigma^{2}_{\Lambda} I )
\end{split}\label{eqn:dpmm_nonconjugate}
\end{align}

\begin{figure*}[ht!]
	\centering
	\includegraphics[width=.95\linewidth]{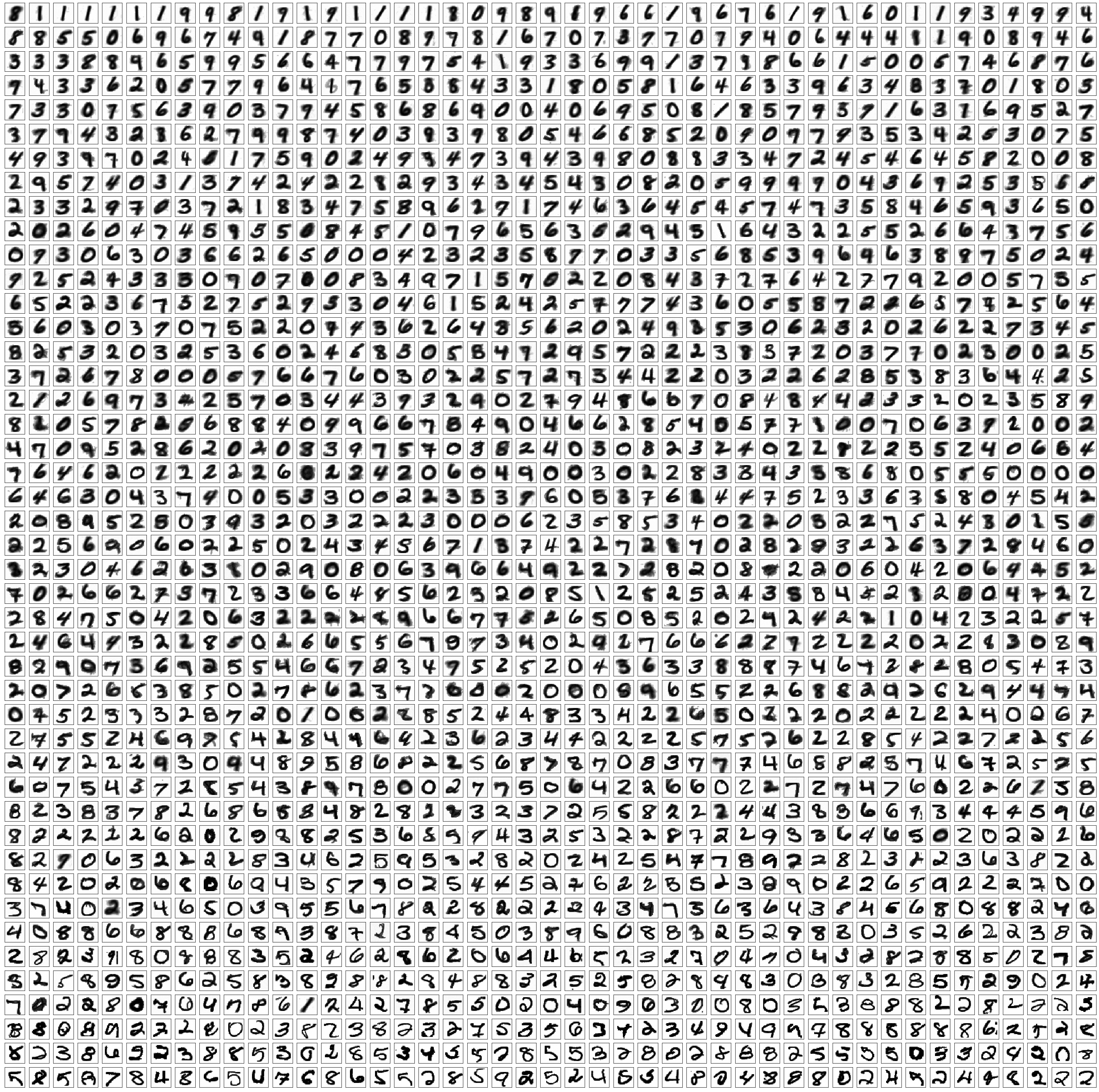}
	\caption{MNIST features obtained via accelerated sampling, sorted in descending order of popularity for the multinomial-log normal model.}
	\label{fig:mnist_acc_feat_exp_norm}
\end{figure*}

\begin{figure*}[ht!]
	\centering
	\includegraphics[width=1.0\linewidth]{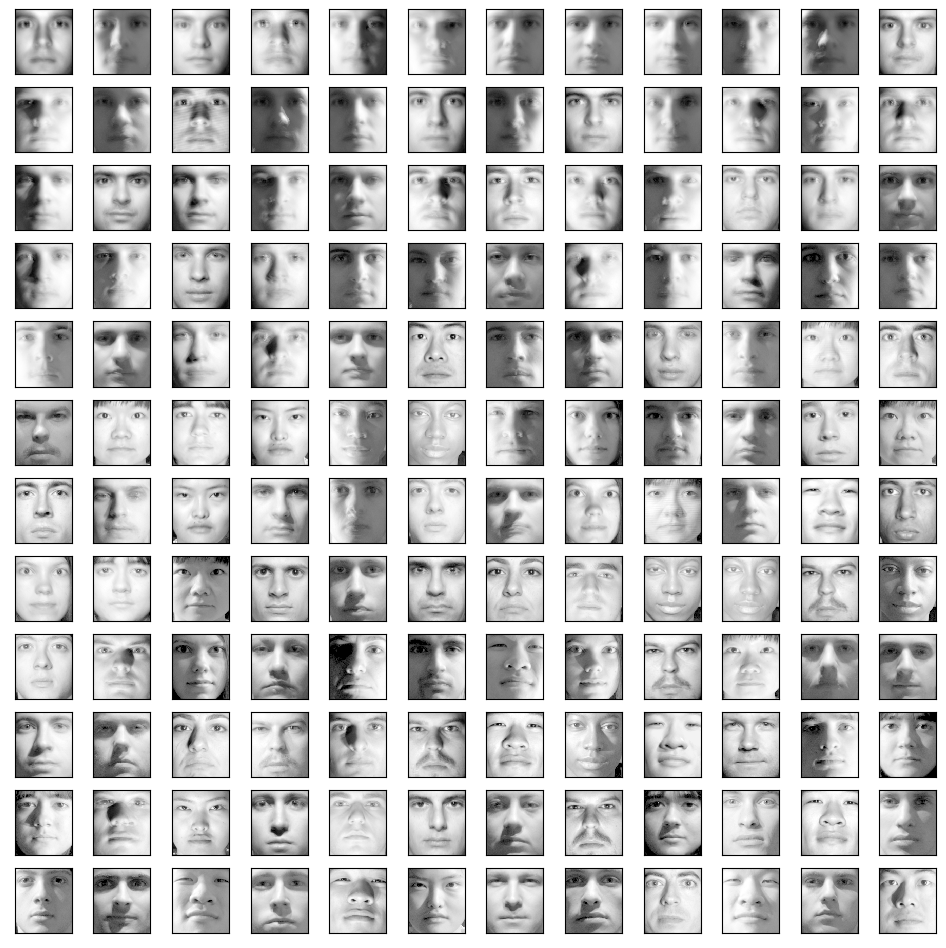}
	\caption{Yale faces features obtained via accelerated sampling, sorted in descending order of popularity for the multinomial-log normal model.}
	\label{fig:yale_acc_feat_exp_norm}
\end{figure*}

\begin{figure*}[ht!]
	\centering
	\includegraphics[width=1.0\linewidth]{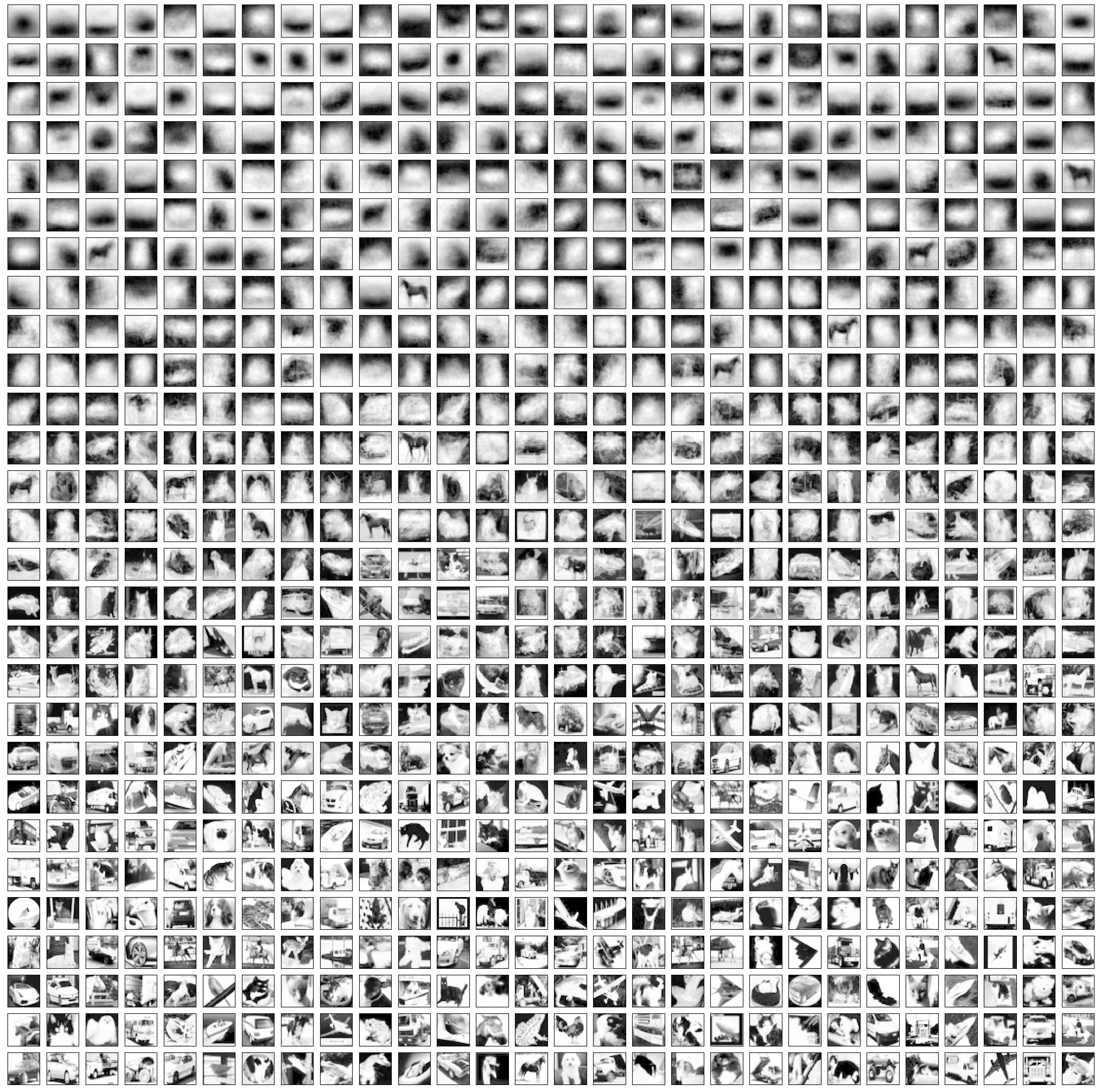}
	\caption{CIFAR features obtained via accelerated sampling, sorted in descending order of popularity for the multinomial-log normal model.}
	\label{fig:cifar_acc_feat_exp_norm}
\end{figure*}

We use the same experimental settings for the duration of the experiments and sampling $\alpha$ as in the previous image examples. For posterior inference of $ \mu_k $ and $ \Lambda_{r} $, we use the elliptical slice sampler \citep{Murray:Adams:MacKay:2010}. Our experimental results show that we can recover clusters that represent meaningful representations of the data, as seen in Figures~\ref{fig:mnist_acc_feat_exp_norm}, \ref{fig:yale_acc_feat_exp_norm}, \ref{fig:cifar_acc_feat_exp_norm}. To construct inference for such a model using variational inference, we would have to derive non-trivial results of ELBO-minimizing variational parameter updates, or resort to computationally intensive methods for variational updates. Here, the only non-trivial change to the earlier model is taking samples of the posterior for $ \mu_k $ and $ \Lambda $, for which there exist a vast library of effective posterior samplers exist. In our setting, because we assume functionals of Gaussian priors, we can use techniques that are fairly simple to use like the elliptical slice sampler. However, even with non-Gaussian priors, we can still take advantage of popular posterior inference algorithms like Hamiltonian Monte Carlo as long as we can take gradients of the log joint distribution of the likelihood and prior (which covers a wide class of likelihood-prior pairs).



\section{Discussion}\label{sec:discussion}
We have introduced a novel method to overcome a problem inherent in Bayesian nonparametric latent variable models of learning unobservable features in a high-dimensional regime while also providing a data-parallel inference method suitable for ``big data'' scenarios. In order to accelerate the mixing of the MCMC sampler, we propose feature locations from observations that are poorly fit to its currently allocated feature during the accelerated stage of our sampler in order to find better locations for features. 

After running accelerated sampling, our method then reverts to an exact algorithm, which is easily and inherently parallelizable without excessive processor communication, in order to maintain the theoretical correctness of our inference algorithm converging to the correct limiting posterior distribution. Although the experiments in this paper only consider Dirichlet proccess examples, our method is easily extendable to other BNP priors for mixture or admixture models like the Pitman-Yor process or hierarchical Dirichlet processes using the same parallelization strategy described in \cite{Zhang:2016}. 

At first, it may seem that the number of features discovered for the accelerated method is excessive but we are using a very simple likelihood to model the digits instead of a more complicated model that is invariant to rotations or scalings of the data. This is apparent in the features discovered for accelerated sampling and variational inference, where a large proportion of the popular clusters learned are various forms of ``ones'' in the case of the MNIST data set. Furthermore, we could propose merge steps for features in order to prune the number of features if we are concerned about the number of features learned. However, we have shown that the additional benefit of our sampler is that our technique can work for a general choice of likelihood and prior, whereas using a collapsed sampler limits the model choice to a narrow range of options for data modeling. Our flexibility in model choice hopefully opens up an opportunity to effectively use more appropriate choices of priors and likelihoods for Bayesian non-parametric problems.


In future work we hope to extend our accelerated sampler to nonparametric latent factor models, like with Indian buffet process-based models. However, it is less obvious how to propose new feature in an accelerated manner for feature allocation models, like in IBP-based models. One possible data-driven proposal mechanism we could use here is to propose new features based on the residuals of the observed data with its instantiated features. We could propose new features based on which observations have the largest residual error, which suggests that there are remaining uninstantiated features which could be present in the data. Moreover, we hope to demonstrate the continued success of Bayesian nonparametrics in modeling complex data while demonstrating the additional benefit that Bayesian and Bayesian nonparametric methods have in providing a natural representation of uncertainty quantification of our results and predictions while having a theoretically motivated methodology of adapting model complexity to the data.


\clearpage
\bibliographystyle{apalike}
\bibliography{acceleratedDP}
\clearpage
\section*{Appendix}

\begin{figure*}[ht!]
	\centering
	\includegraphics[width=.95\linewidth]{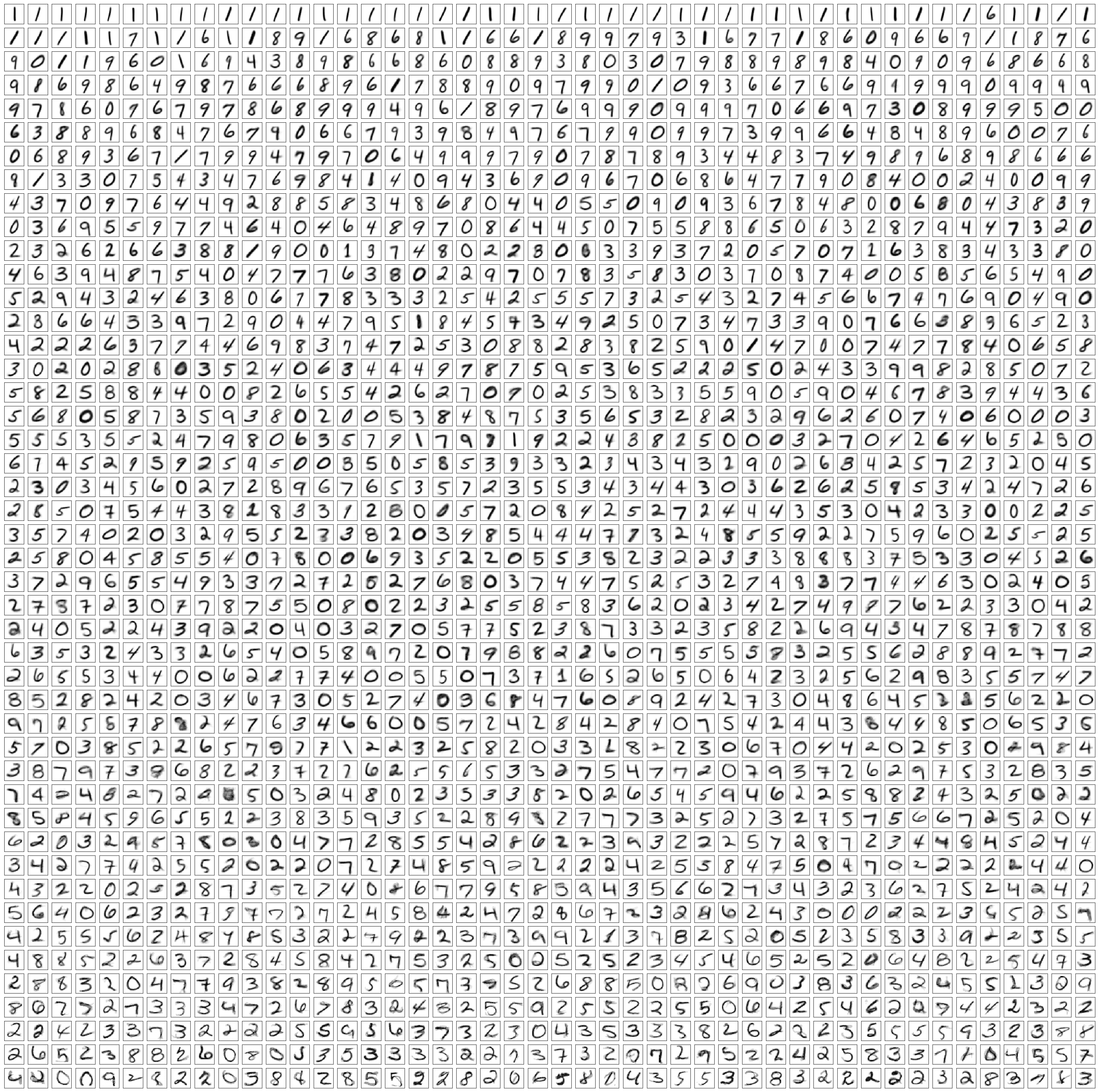}
	\caption{MNIST features obtained via accelerated sampling, sorted in descending order of popularity for the multinomial-Dirichlet model.}
	\label{fig:mnist_acc_feat}
\end{figure*}

\begin{figure*}[ht!]
	\centering
	\includegraphics[width=1.0\linewidth]{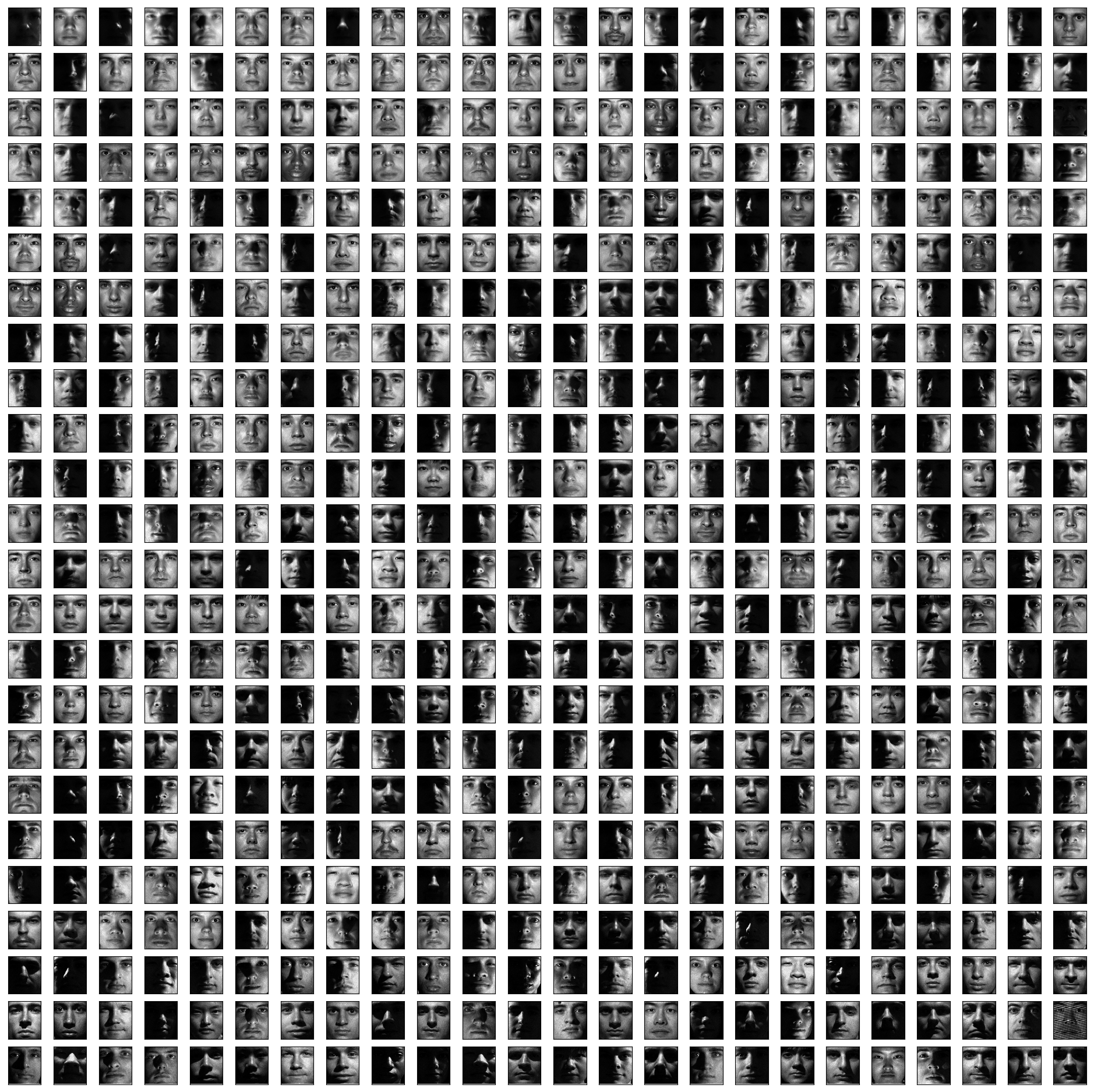}
	\caption{Yale faces features obtained via accelerated sampling, sorted in descending order of popularity for the multinomial-Dirichlet model.}
	\label{fig:yale_acc_feat}
\end{figure*}

\begin{figure*}[ht!]
	\centering
	\includegraphics[width=1.0\linewidth]{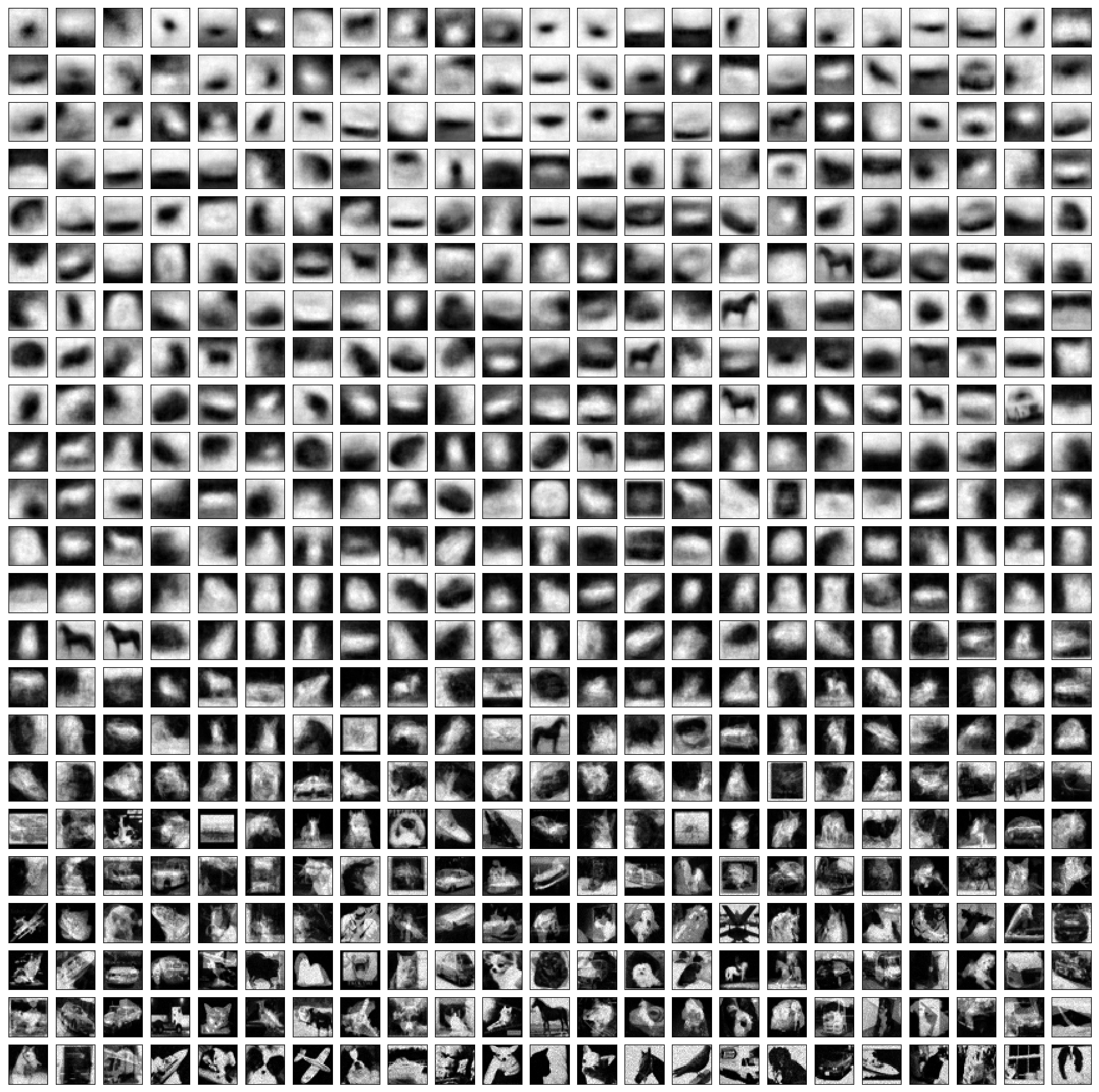}
	\caption{CIFAR features obtained via accelerated sampling, sorted in descending order of popularity for the multinomial-Dirichlet model.}
	\label{fig:cifar_acc_feat}
\end{figure*}

\begin{figure*}[ht!]
	\centering
	\includegraphics[width=1.0\linewidth]{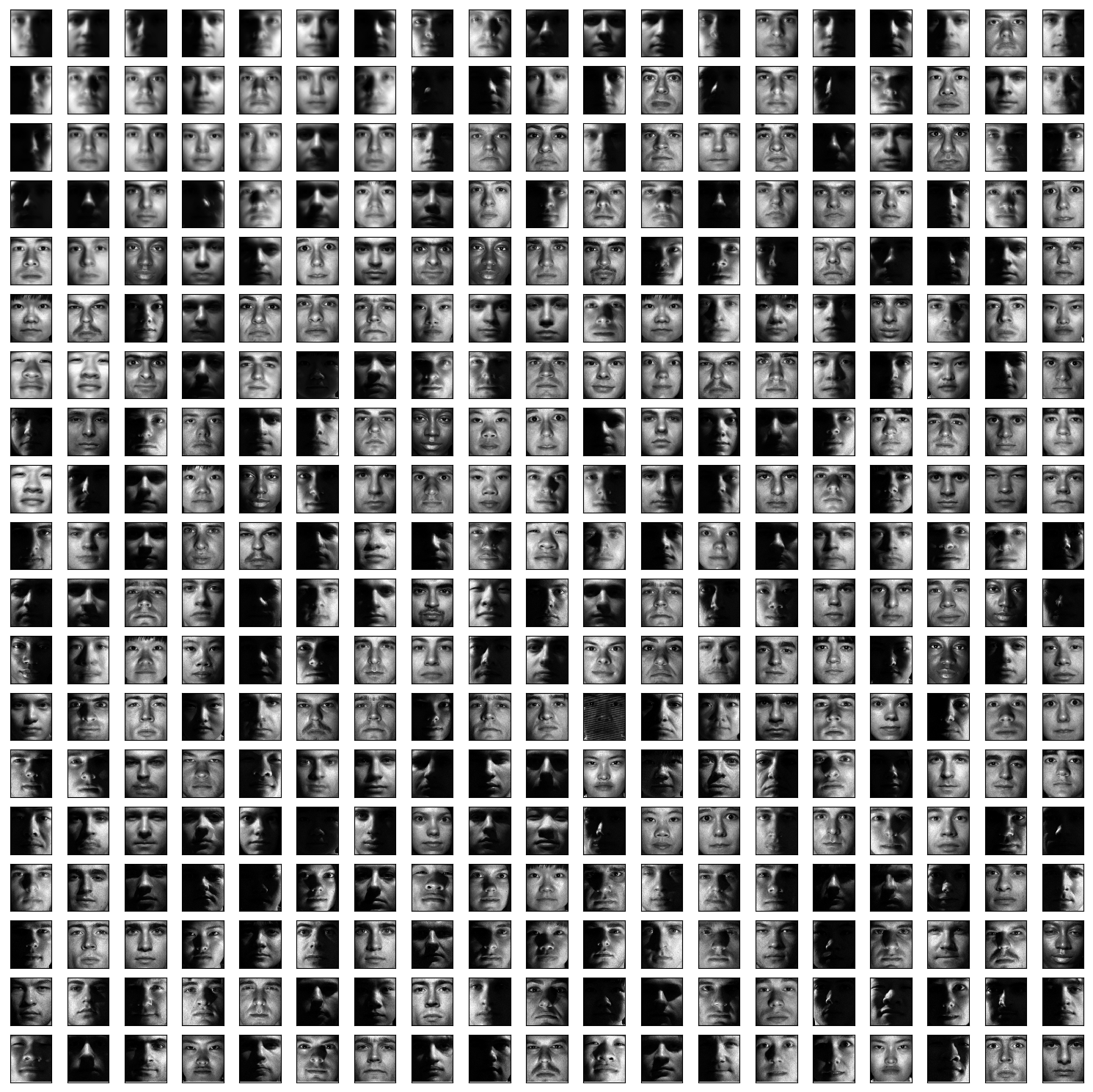}\\
	\caption{Yale faces features  obtained via variational inference, sorted in descending order of popularity for the multinomial-Dirichlet model.}
	\label{fig:yale_feat_collapsed}
\end{figure*}

\begin{figure*}[ht!]
	\centering
	\includegraphics[width=1.0\linewidth]{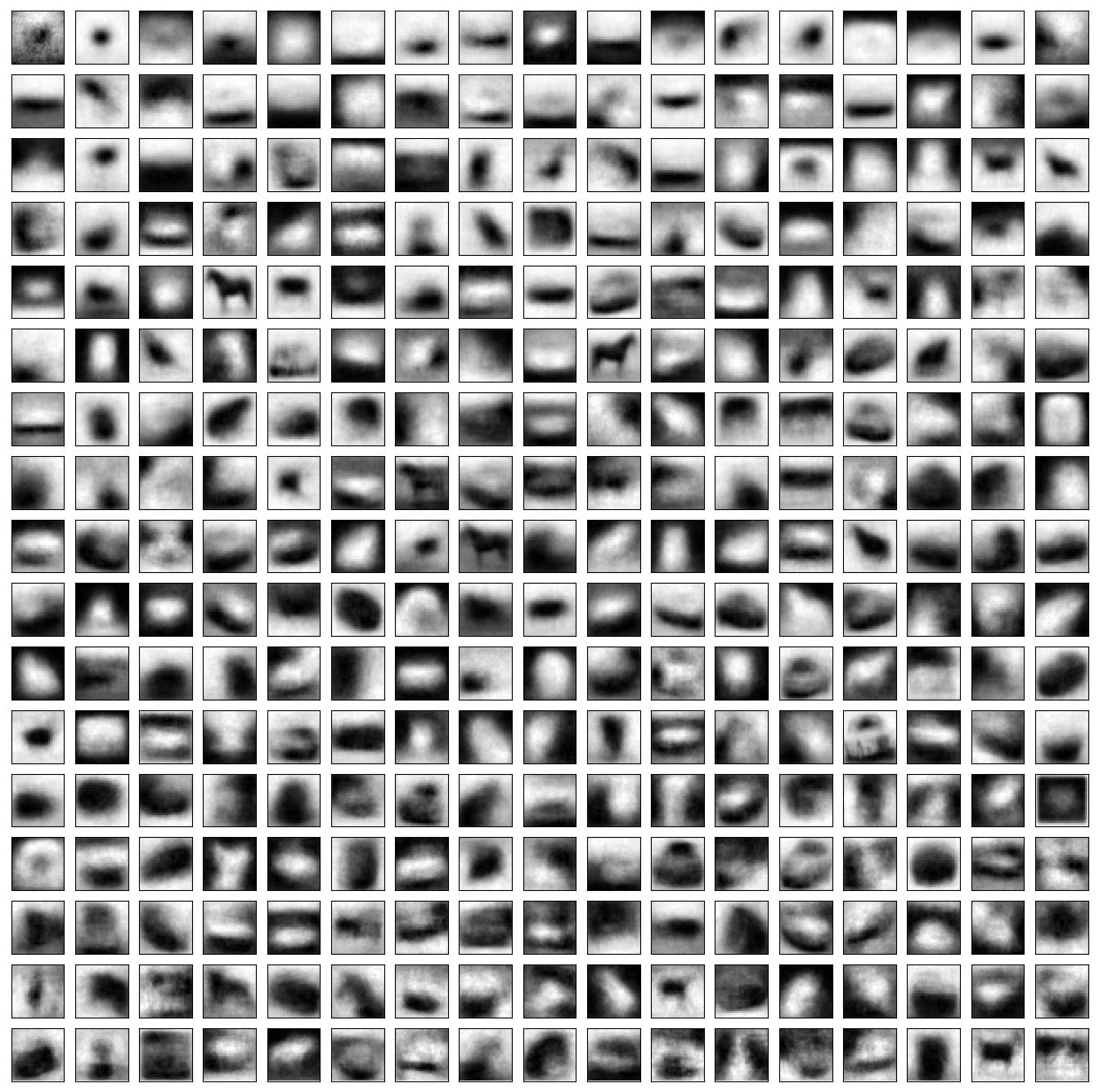}\\
	\caption{CIFAR faces features obtained via collapsed sampling, sorted in descending order of popularity for the multinomial-Dirichlet model.}
	\label{fig:cifar_feat_collapsed}
\end{figure*}

\end{document}